%% file: acl2023.tex
\pdfoutput=1

\documentclass[11pt]{article}

\usepackage{ACL2023}

\usepackage{times}
\usepackage{latexsym}

\usepackage[T1]{fontenc}

\usepackage[utf8]{inputenc}

\usepackage{microtype}

\usepackage{inconsolata}

\usepackage{amssymb}
\usepackage{amsmath}
\usepackage{tabularx}
\usepackage{multirow}
\usepackage{caption}
\usepackage{subcaption}
\usepackage[capitalize]{cleveref}
\crefname{equation}{Eq.}{Eqs.}
\crefname{section}{Sec.}{Sec.}
\Crefname{section}{Section}{Sections}
\crefname{figure}{Fig.}{Figs.}
\Crefname{figure}{Figure}{Figures}
\crefname{appendix}{App.}{Apps.}
\Crefname{appendix}{Appendix}{Appendices}
\crefname{table}{Tab.}{Tabs.}
\Crefname{table}{Table}{Tables}
\usepackage{booktabs}       %
\usepackage{diagbox}
\usepackage{xspace}

\usepackage[frozencache,cachedir=.]{minted}
\usepackage{algorithm}
\floatname{algorithm}{Alg.}
\crefname{algorithm}{Alg.}{Algs.}
\Crefname{algorithm}{Algorithm}{Algorithms}

\usepackage{hyperref}
\usepackage{xcolor}         %
\usepackage[normalem]{ulem}

\usepackage{nicefrac}
\usepackage{graphicx}
\graphicspath{ {./graphs/} }
\usepackage{pgf}

\newcommand{\newparagraph}[1]{\noindent {\bf #1}}
\renewcommand{\newparagraph}[1]{\paragraph{#1}}
\newcommand{\com}[1]{}
\newcommand{\resolved}[1]{}

\newcommand{\msrnn}[0]{MSRNN\xspace}
\newcommand{\msrnns}[0]{MSRNNs\xspace}
\newcommand{\method}[0]{TOVA\xspace}
\newcommand{\window}[0]{Window\xspace}
\newcommand{\windowi}[1]{Window\(+#1\)\xspace}
\newcommand{\hto}[0]{H\(_2\)O\xspace}
\newcommand{\finite}[0]{bounded\xspace}
\newcommand{\infinite}[0]{unbounded\xspace}
\newcommand{\Finite}[0]{Bounded\xspace}
\newcommand{\Infinite}[0]{Unbounded\xspace}

\newcommand{\draftcomment}[3]{{\textcolor{#3}{[#1]#2}}}
\newcommand{\roy}[1]{\draftcomment{#1}{\textsc{Roy}}{pink}}
\newcommand{\michael}[1]{\draftcomment{#1}{\textsc{Michael}}{blue}}
\newcommand{\matanel}[1]{\draftcomment{#1}{\textsc{Matanel}}{teal}}
\newcommand\addtag{\refstepcounter{equation}\tag{\theequation}}
\newcommand{\rs}[1]{\roy{\sout{#1}}}
\newcommand{\ra}[1]{\textcolor{pink}{[#1]}}
\newcommand{\rr}[2]{\rs{#1}\ra{#2}}

\newcommand{\acl}[1]{}

\newcommand{\arxiv}[1]{#1}

\newcommand {\R}{\mathbb{R}}

\title{Transformers are Multi-State RNNs}

\author{
\textbf{Matanel Oren}\thanks{~~Equal contribuation}\(^{*, H}\) \quad 
\textbf{Michael Hassid}\(^{*, H, M}\) \quad 
\textbf{Nir Yarden}\(^{H}\) \\ \\ %
\textbf{Yossi Adi}\(^{H, M}\) \quad
\textbf{Roy Schwartz}\(^{H}\) \quad \AND
\normalfont{$^H$The Hebrew University of Jerusalem} \quad
\normalfont{$^M$FAIR, AI at Meta}
\\ \\
{\tt \{matanel.oren,michael.hassid\}@mail.huji.ac.il}
}

\begin{document}

\maketitle
\begin{abstract}

\input{0_abstract}

\end{abstract}

\section{Introduction}
\label{sec:introduction}

\input{1_introduction}

\section{Background}
\label{sec:background}
\input{2_background}

\section{Transformers as Multi-State RNNs}
\label{sec:method}
\input{3_method}

\section{Experimental Setup}
\label{sec:exp_setup}
\input{4_exp_setup}

\section{Results: Pretrained Transformers Often Act as Bounded MSRNNs}%
\label{sec:results}

\input{5_results}

\section{Analysis }
\label{sec:analysis}
\input{6_analysis}

\section{Related Work}
\label{sec:related_work}
\input{7_related_work}

\section{Conclusion}
In this work, we redefined decoder transformers as a form of multi-state RNNs~(\msrnn) with an \infinite multi-state size. We then showed that they can be compressed to \finite \msrnns by limiting the number of tokens they can handle at each decoding step. 

We introduced \method, a conceptually simple compression method that selects which tokens to keep using their attention scores. 
We showed that \method is superior compared to existing compression policies; in many cases, \method performs comparably to the full (\infinite) model, while requiring \nicefrac{1}{8}--\nicefrac{1}{4} of the multi-state size. 
\method also allows processing long inputs, up to 70K tokens. %

Our findings shed light on the inter-working of transformers, and their connections to RNNs. They also have practical value---they can reduce the LLM cache size by up to 88\% and increase throughput by 4.8X.

\section*{Limitations}
Evaluating models on long text generation is computationally expensive and might limit others from reproducing our results. 
Further, the evaluation of such task is extremely complicated, even for humans. We therefore resort to GPT-4 to compare the output of our \method policy compared to the topline model~(\cref{subsec:text_gen}). We recognize that this is far from perfect, and will most likely not catch the full breadth of evaluating text quality. 
Finally, our evaluation framework focuses on English tasks. It is not unlikely that languages with more flexible word order will make different use of the attention mechanism, and thus might require a larger multi-state size.

\section*{Ethics Statement}
Our work has the potential to dramatically reduce the memory footprint of transformer LLMs, thereby potentially increasing their adoption by users with limited hardware access.

This work does not collect any new data, and only uses open source models, and public data collected by other sources.

\arxiv{
\section*{Acknowledgements}
\input{acknowledgements}
}
\bibliography{anthology,custom}
\bibliographystyle{acl_natbib}

\newpage

\appendix
\section{Policy Ablation}
\label{sec:abalation_app}
\input{app_ablation}
\input{app_algo}

\section{Prompts}
\label{sec:prompts_app}
\input{app_prompts}

\section{Details of Generation Evaluation}
\label{sec:generation_app}
\input{generation_app}

\section{Experimental Details}
\label{sec:lm_exp_det}
\input{app_lm_implementation}

\section{Long Range Understanding with Base Models}
\label{sec:leu_app}
\input{app_long_range}

\section{Illustration of the Tokens Retained by \method}\label{sec:extended_analysis_app}
\input{app_ext_analysis}

\section{Full Part-of-Speech Tag Analysis}
\label{sec:pos_app}
\input{app_pos}

\end{document}

%% file: 0_abstract.tex
Transformers are considered conceptually different from the previous generation of state-of-the-art NLP models---recurrent neural networks (RNNs). In this work, we demonstrate that decoder-only transformers can in fact be conceptualized as \infinite multi-state RNNs—an RNN variant with unlimited hidden state size.
We further show that transformers can be converted into \textit{\finite} multi-state RNNs by fixing the size of their hidden state, effectively compressing their key-value cache. 
We introduce a novel, training-free compression policy---\textbf{T}oken \textbf{O}mission \textbf{V}ia \textbf{A}ttention~(\method).\arxiv{\footnote{Literally ``good'' in Hebrew.}}
Our experiments with four long range tasks and several LLMs show that \method outperforms several baseline compression policies. Particularly, our results are nearly on par with the full model, using in some cases only \nicefrac{1}{8} of the original cache size, which translates to 4.8X higher throughput. 
Our results shed light on the connection between transformers and RNNs, and help mitigate one of LLMs' most painful computational bottlenecks---the size of their key-value cache.%
\arxiv{\footnote{\url{https://github.com/schwartz-lab-NLP/TOVA}}}

%% file: 1_introduction.tex
Not so long ago, transformers~\cite{vaswani2017attention} replaced recurrent neural networks~(RNNs;~\citealp{ELMAN1990179}) as the go-to architecture for NLP. Transformers are considered conceptually different than RNNs; they have direct access to each token representation in the sequence, while RNNs maintain a recurring state of previous inputs. 
Recently, \textit{decoders} became a dominant transformer variant for large language models (LLMs;~\citealp{brown2020language, touvron2023llama, jiang2023mistral}).
These typically generate their output autoregressively---the generation of each token representation depends on the key and value computation of previous~tokens.\footnote{These previous computations are often cached for efficiency purposes, referred to as KV caching~\cite{radford2019language, pope2022efficiently}. We note that the arguments we make in this work apply similarly to non-cached implementations.}

\begin{figure}[t]
\centering
\includegraphics[trim={0.1cm 8.1cm 0.1cm 0cm}, clip, width=\columnwidth]{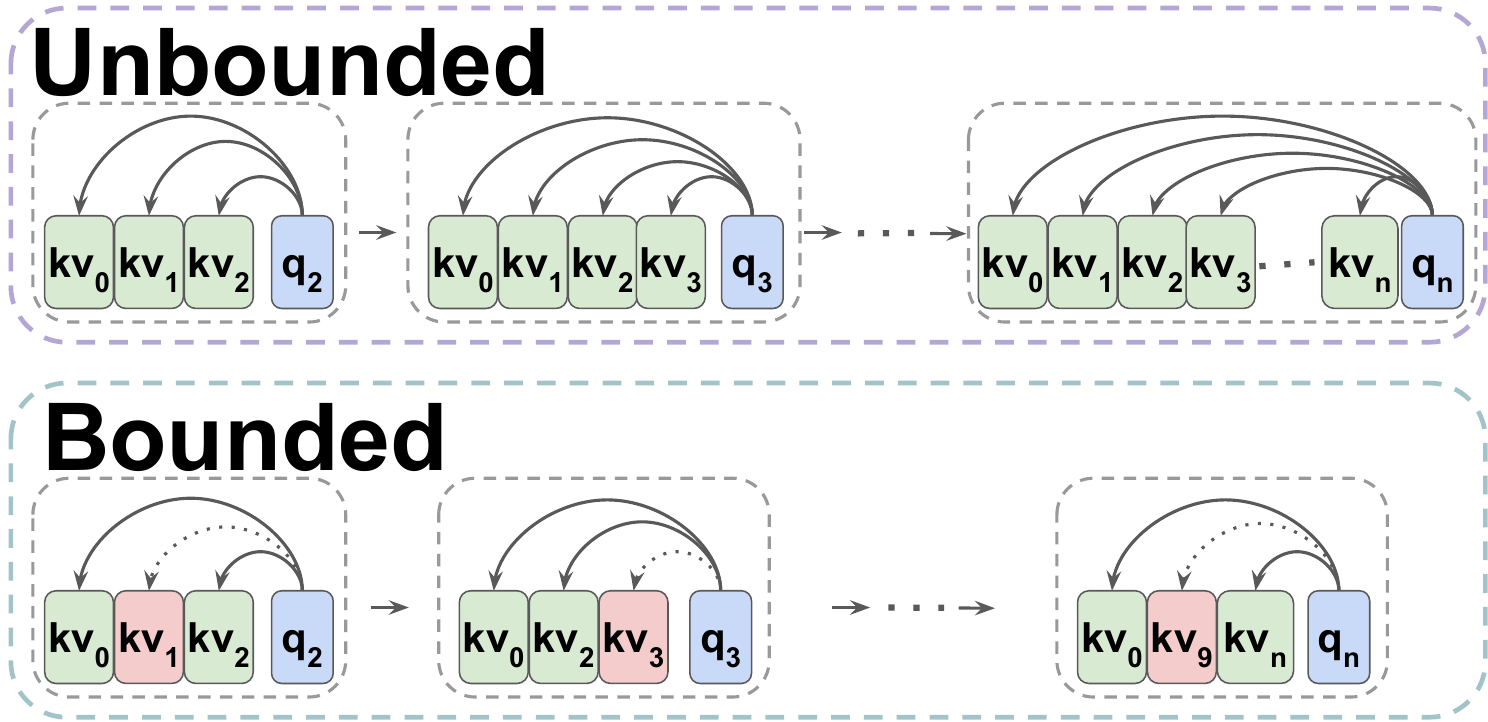}

\caption{Top: transformers can be thought of as \infinite multi-state RNNs~(\msrnns), with the key-value vectors corresponding to a multi-state that dynamically grows infinitely~({\color{YellowGreen}{green}} elements). Bottom: transformers can be converted to \textit{\finite} \msrnns, which keep a fixed-size multi-state (here of size 2), by dropping one state ({\color{Salmon}{red}} elements) at each decoding step.
\label{fig:fig1}}
\end{figure}

In this work, we demonstrate that the autoregressivity of transformers aligns with the core principle of RNNs---preserving a state from one step to the other. We formally redefine decoder-only transformers as \textit{multi-state} RNNs~(\msrnn)---a generalized version of RNNs with multiple states, each corresponding to a history token.
Importantly, as the number of tokens grows with each decoding step, transformers correspond to \msrnns with an \textit{\infinite} number of states~(\cref{fig:fig1}, top).

We then show that transformers can be compressed into \textit{\finite} \msrnns by limiting their number of states~(\cref{fig:fig1}, bottom).
This process requires a compression policy for selecting the states to retain. While existing methods, e.g., windowed attention~\citep{wang-etal-2019-multi}, can be cast as such policies, we propose a novel policy, \method, which retains the states with the highest attention~scores.

We experiment with four long range tasks, several leading LLMs, and a few baseline compression policies. Our results show that \method outperforms all baselines in all setups. Further, using \method can match the performance of the full~(uncompressed) model using as little as \nicefrac{1}{8} of the full model multi-state, which leads to a throughput increase of up to 4.8X. Finally, \method allows running on dramatically longer contexts, up to 70K tokens.

We finish by analyzing the states kept in memory by \method, and the tokens they correspond to. Unlike previous work~\citep{streamllm,hto}, we observe that not all recent tokens are important to retain, and some may be safely dropped. We also show the importance of keeping the very first token in the sequence, as well as other, perhaps surprising tokens like possessive endings.

Our findings shed light on the connection between transformers and RNNs. They also help mitigate the LLM memory bottleneck during decoding, which directly translates to higher throughput.

%% file: 2_background.tex
\subsection{RNNs}
\label{subsec:background_rnn}
Recurrent Neural Networks (RNNs;~\citealp{ELMAN1990179})
process sequential data recurrently. In the most general form, each layer \(l\) (often called a \textit{cell}) is modeled as a function \(f_{\text{\tiny{RNN}}}^l\) that receives at time \(t\) two inputs: \(x_t^l\), a representation of the current token, and \(h_{t-1}^l\), the hidden state from the previous step. It then outputs two values: \(x_t^{l+1}\), an updated token representation, and \(h_t^l\), a new hidden state:
\[
x_t^{l+1}, h_t^l = f_{\text{\tiny{RNN}}}^l(x_t^l, h_{t-1}^l) \label{eq:rnn_def} \addtag
\]
\(h_t^l\) is used for the recurrent computation over the next token \(x_{t+1}^l\),
while \(x_t^{l+1}\) is used as input to the next layer. It is common, though not necessary, to set \(x_{t}^{l+1} := h_t^l\), i.e., the input for the following layer and the hidden state are the same. 

\subsection{Transformers}
\label{subsec:background_trans}
Transformers~\cite{vaswani2017attention} process sequential data non-recurrently. A transformer layer \(f_{\text{\tiny{TRANS}}}^l\) takes as input a sequence of token representations of hidden size $d$: \(X^l = (x_1^l, ..., x_t^l)^T \in \R^{t \times d}\) and returns a transformed representation:
\[
X^{l+1} = f_{\text{\tiny{TRANS}}}^l(X^l) = \text{FF}^l\bigl(\text{SelfAttn}^l(X^l) \bigr) \addtag
\]
Each transformer layer consists of two main components: self-attention~(SelfAttn\(^l\)) and Feed-Forward~(FF\(^l\)).\footnote{Layer normalization, skip connections, and multiple attention heads are omitted for brevity.} The former operates over the entire sequence, while the latter on each token individually. Self-attention projects the input into three matrices: \(Q^l, K^l, V^l \in \R^{t \times d}\), and computes:
\begin{align}
X^l_{attn} & = \text{Attn}(Q^l, K^l, V^l) \\
& = \underset{A^l}{\underbrace{\text{Softmax}\bigl(Q^l \cdot (K^l)^T\bigr)}} \cdot V
\end{align}
where \(A^l \in \R^{t \times t}\), the attention matrix, computes the interactions between tokens within a sequence. 

In this work we focus on transformer \textbf{decoders}, which mask the upper triangular part of the attention matrix to perform next-token prediction.
During decoding, it is common to cache the \(K, V\) matrices to avoid recomputing previous tokens.

%% file: 3_method.tex
We start by formally defining a new RNN variant, Multi-State RNN~(\msrnn;~\cref{subsec:MSRNN}). We then show that transformers can be viewed as \msrnns with an \infinite number of states~(\cref{subsec:trans_msrnn}), and that their number of states can be bounded by applying a compression policy~(\cref{subsec:MSRNN_cap}). We finish by discussing LLMs as \msrnns~(\cref{subsec:LLM_MSRNN}).

\subsection{Multi-State RNNs}
\label{subsec:MSRNN}
We define an \msrnn as an RNN with a state~\textit{matrix} instead of a vector:~\(H_t^l \in \R^{g(t) \times d}\). 
The \msrnn equation corresponding to \cref{eq:rnn_def} is:
\[
x_t^{l+1}, H_t^l = f_{\text{\tiny{\msrnn}}}^l(x_t^l, H_{t-1}^l) \label{eq:msrnn_def} \addtag
\]
We can interpret each row of \(H_t^l\) as a single-state, allowing us to think of \(H_t^l\) as a multi-state matrix.\footnote{We could unroll the matrix and define it as a single vector in \(\R^{g(t) \cdot d}\) and use the traditional RNN terminology, but we find it more convenient to think of it as a matrix.}

The size~of $H_t^l$ is parameterized by a function \(g\). Setting \(g(t)=1\) for all \(t\) reduces an \msrnn to a standard (single-state) RNN. Setting \(g(t) \le k\) for a constant \(k\) restricts it to a \textit{\finite} memory capacity. If \(g\) is unbounded in \(t\), the \msrnn state can have \textit{\infinite} capacity. 

\subsection{Transformers are \Infinite \msrnns}
\label{subsec:trans_msrnn}

Consider the case where \(g(t)=t\), i.e., the number of states equals the number of input tokens in the current time-step.
In this setup, we~can view a transformer as an \infinite \msrnn, where \(H_t^l=(K_t^l, V_t^l)\) and the layer computation is:
\begin{align}
(K_t^l, V_t^l) & = \Bigl(\Bigl(\substack{K_{t-1}^l \\ k_t^l}\Bigr), \Bigl(\substack{V_{t-1}^l \\ v_t^l}\Bigr) \Bigr) \label{eq:trans_as_msrnn} \\
x_t^{l+1} & = \text{FF}^l\bigl(\text{Attn}^l(q_t^l, K_t^l, V_t^l) \bigr)   
\end{align}
where \(q_t^l, k_t^l, v_t^l\) are the self-attention projections of \(x_t^l\), and each state of \((K_t^l,V_t^l)\) corresponds to a specific token. Combined, we get the \msrnn equation for transformers:
\[
x_t^{l+1}, (K_t^l, V_t^l) = f_{\text{\tiny{TRANS}}}^l\bigl(x_t^l, (K_{t-1}^l, V_{t-1}^l) \bigr) \addtag
\]

\subsection{Converting Transformers into \Finite \msrnns}
\label{subsec:MSRNN_cap}
Transformers can be  converted into \finite \msrnns by setting \(g(t) = \text{min}(t,k)\) for some~\(k\). When \(t\) exceeds \(k\), a compression policy should be applied in order to fit the multi-state to into the bounded memory.

Interestingly, several existing KV cache compression  methods, e.g., windowed attention~\cite{wang-etal-2019-multi} and \hto~\cite{hto}, can be seen as such compression policies, see~\cref{subsec:baselines}. %

\subsection{LLMs as \msrnns}
\label{subsec:LLM_MSRNN}
LLMs are generally built as transformer decoders.
As such, they are, on the one hand, unbounded \msrnns~(\cref{subsec:trans_msrnn}). On the other, they are trained~with a fixed context length, and often struggle at extrapolating beyond it~\cite{alibi}, and thus may be considered \finite.

We argue that LLMs are indeed \textit{\infinite}: At inference time, they can process any number of tokens, and are limited only by the available
memory. In addition, both at training and inference time, they accumulate token representations into their multi-state without dropping any from their memory. Thus, as memory compression is the fundamental feature of \finite \msrnns, LLMs should be conceptualized as \infinite. Interestingly, we later show that despite their \infinite capacity, they often act in practice as \finite \msrnns.

\section{\method: Token Omission Via
Attention}\label{sec:tova}

\begin{figure}[t]
\centering
\includegraphics[trim={0.15cm 3.8cm 5.5cm 0.6cm}, clip, width=0.7\columnwidth]{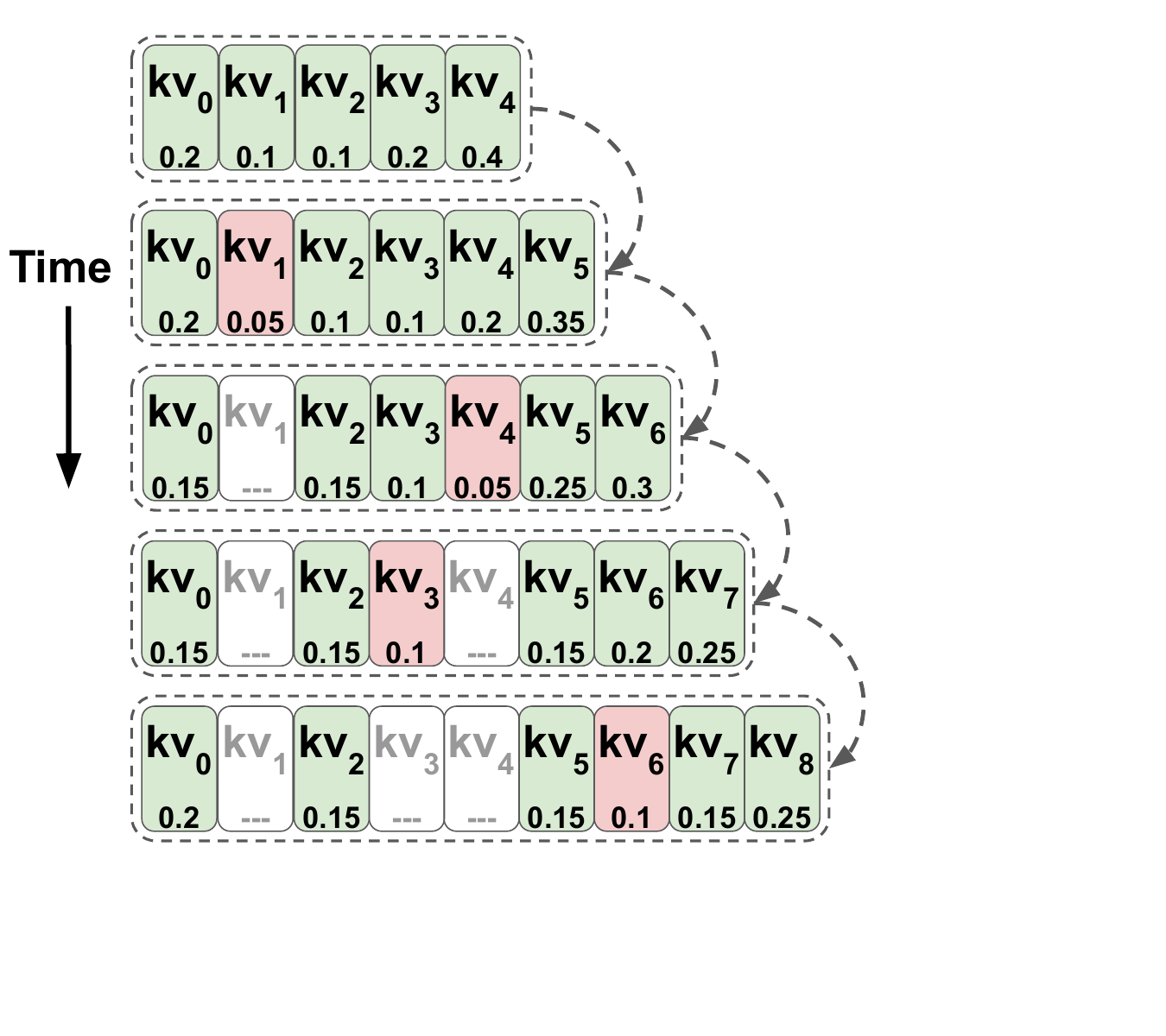}

\caption{
\method 
policy keeps a fixed-size multi-state ({\color{YellowGreen}{green}} cells). At each decoding step (different rows), the state with the lowest attention score is omitted~({\color{Salmon}{red}} cells, which become 
{\color{lightgray}{transparent}} in subsequent steps).
\label{fig:fig2}}
\end{figure}

Converting an \infinite \msrnn to a \finite one requires a state-compression policy~(\cref{subsec:MSRNN_cap}).
We introduce \method---a novel, training-free policy for doing so~(\cref{fig:fig2}). After the multi-state reaches the capacity limit, \method drops at each decoding step the token with the lowest attention score. %
Formally, when \(t > k\) and assuming \(j\) is the state with the lowest attention score, \method applies the following over the multi-state \( (K_t^l, V_t^l)\) from \cref{eq:trans_as_msrnn}:
\begin{align}
(K_t^l, V_t^l) & = \Bigl(\Bigl(\substack{K_{0:j-1}^l \\ K_{j+1:k}^l}\Bigr), \Bigl(\substack{V_{0:j-1}^l \\ V_{j+1:k}^l}\Bigr) \Bigr)
\end{align}

\method computes the attention scores of each head separately, and can thus retain different tokens at different heads.
In practice, preliminary results show that averaging the attention scores across the heads of a given layer is superior to considering each head individually~(\cref{sec:abalation_app}). %
See~\cref{sec:tova_algo} for a torch-like implementation of \method.

%% file: 4_exp_setup.tex
We aim to check whether transformer LLMs converted into \finite \msrnns can match the performance of the full model (an \infinite \msrnn; \cref{subsec:LLM_MSRNN}). 
Below we describe our baseline compression policies~(\cref{subsec:baselines}), the datasets~(\cref{subsec:eval}), and the LLMs we experiment with~(\cref{subsec:models}).

\subsection{Baseline Compression Policies}
\label{subsec:baselines}
Below we describe previously proposed compression policies. We note that, to the best of our knowledge, we are the first to make the connection between these policies and RNNs. As our focus is on the capacity of off-the-shelf models, we only consider baseline policies that operate on pretrained LLMs and require no additional training. \Cref{sec:related_work} discusses approaches that do require training. 

\newparagraph{\window} This policy~\cite{wang-etal-2019-multi} implements a First In First Out (FIFO) strategy. When the multi-state reaches its capacity, the oldest state~(i.e., the earliest token state) is discarded, such that only the most recent states are kept. 

\newparagraph{\windowi{i}} This policy uses a fixed window, but also retains the first \(i\) states, for some constant \(i\). Previous work~\cite{streamllm, lminfinite} has shown that \windowi{i} strongly outperforms \window using as few as $1$--$4$ early states.

\newparagraph{\hto} Much like \windowi{i}, this policy~\cite{hto} keeps a fixed window of recent tokens, as well as additional earlier tokens. Unlike \windowi{i}, it dynamically selects the non-window tokens by aggregating the attention scores throughout the sequence, and keeping the ones with the highest aggregated scores. 
The number of non-window tokens is typically set as half~of~the multi-state size. Like \method, \hto can operate head-wise or layer-wise. Preliminary results~(\cref{sec:abalation_app}) indicate that both variants perform similarly, so we follow \citet{hto} and use the head-wise~version.
\newparagraph{Full model (topline)}
We use the full (\infinite) model as our topline.
Pretrained transformers struggle with sequences longer than their pretrained sequence length \cite{alibi}. In order to make the most fair comparison, we feed the model with the full training sequence length of the particular LLMs we use, and use smaller multi-state sizes for the different compression policies.\footnote{In \cref{subsec:extra} we also report extrapolation experiments.} 
\newparagraph{}

We note that the all baseline policies presented above introduce strong inductive biases; e.g., devoting a substantial part of the state towards the most recent tokens, and preferring tokens appearing early in the sequence.\footnote{Note that \hto aggregates the attention weights, which favors initial tokens, as they accumulate more attention scores as the sequence progresses.} %
In contrast, \method makes fewer assumptions: it neither fixes a window of recent token-states, nor favors early ones.

\subsection{Long Range Evaluation}
\label{subsec:eval}

To trigger the different policies, we focus on long range evaluation. 
We employ three types of long-range evaluation: language modeling, long-range understanding, and text generation. 
See \cref{sec:prompts_app} for the prompts used for the different tasks.

\newparagraph{Language modeling} We report perplexity on the PG-19 test set~\cite{Rae2020Compressive}, a widely used benchmark for evaluating long range language models~\cite{so2022primer,hutchins2022blockrecurrent, chen2023extending}. PG-19 is composed of 100 full-length books of average length of 70k tokens.

\newparagraph{Long range understanding} We consider two tasks from ZeroSCROLLS~\cite{zeroscrolls}, each focusing on a different aspect of long range understanding:
(a) SQuALITY~\cite{squality}, a question focused summarization dataset; 
and (b) QASPER~\cite{qasper}, a QA dataset based on the S2ORC dataset~\cite{lo-etal-2020-s2orc}. QASPER can be considered a retrieval task, as answering its questions requires retrieving specific details from long texts.
For SQaULITY, we report the geometric mean of ROUGE-1/2/L scores~(based on the gold summary, see~\citealp{zeroscrolls}). For QASPER, we follow~\citet{qasper} and report F1 score.

\begin{figure*}[t!]
     \centering
     \begin{subfigure}[b]{0.32\textwidth}
         \centering
         \includegraphics[trim={0 .65cm 0 .7cm},clip,width=\textwidth]{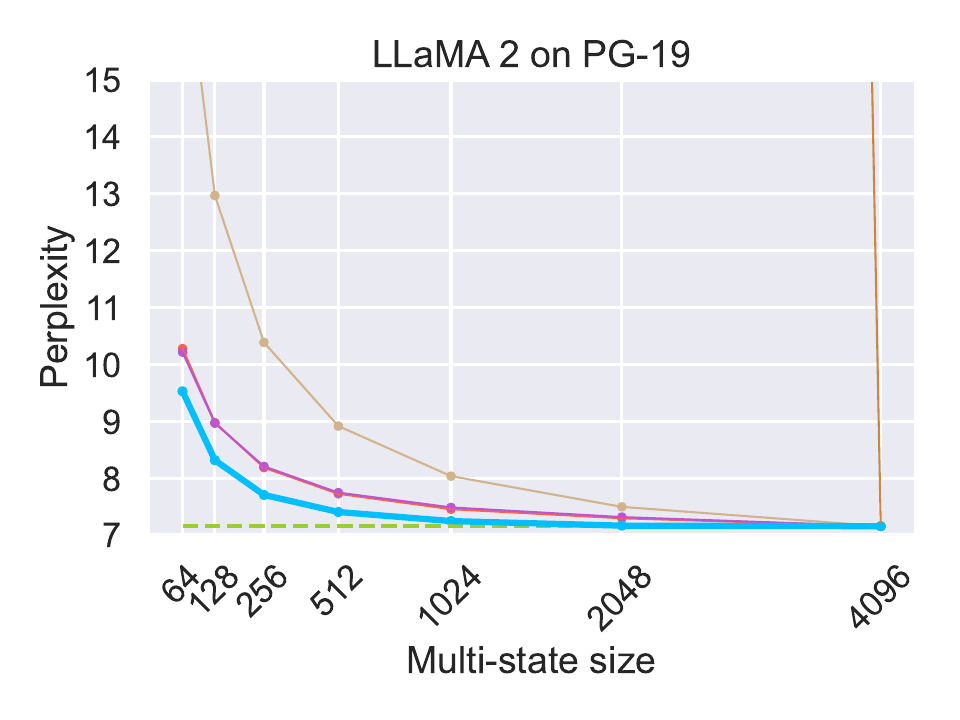}        
     \end{subfigure}
     \hfill
     \begin{subfigure}[b]{0.32\textwidth}
         \centering
         \includegraphics[trim={0 .65cm 0 .7cm},clip,width=\textwidth]{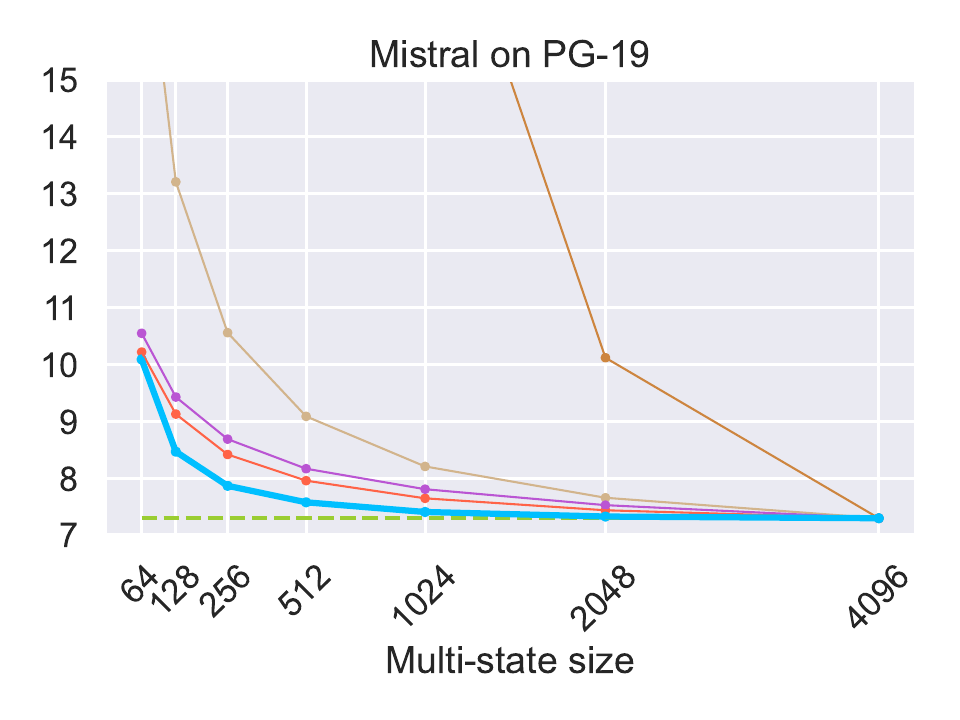}     
     \end{subfigure}
     \hfill
     \begin{subfigure}[b]{0.32\textwidth}
         \centering
         \includegraphics[trim={0 .65cm 0 .7cm},clip,width=\textwidth]{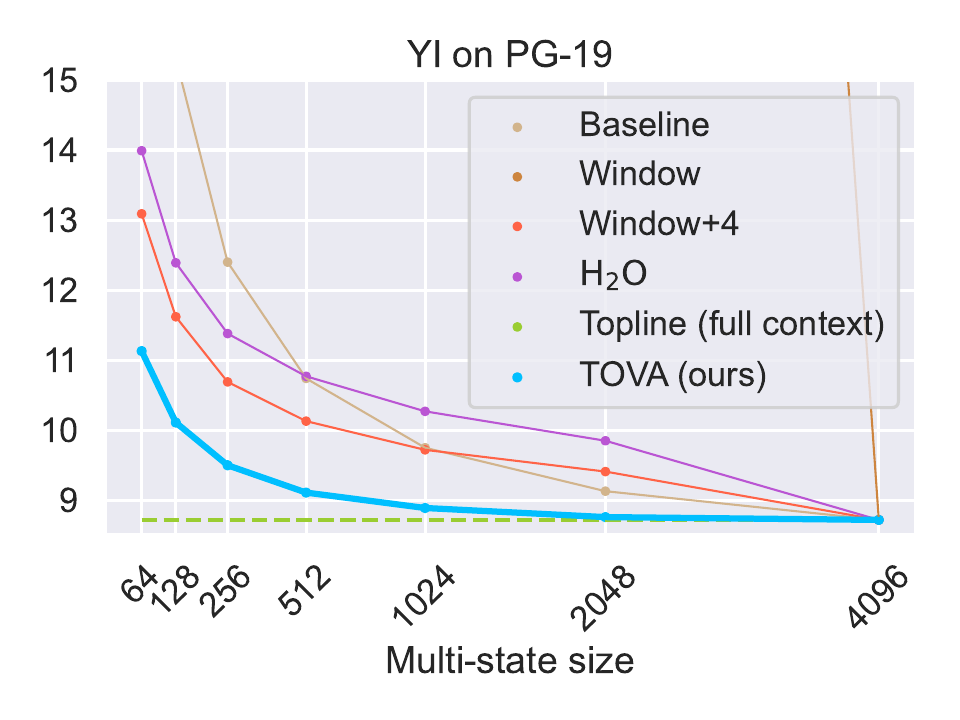}
     \end{subfigure}
     \caption{
     Perplexity results for the PG-19 test set. \method outperforms all other policies in all multi-state sizes, while maintaining comparable results to the full context topline using \nicefrac{1}{8} of the context size.
 \label{fig:ppl}}
\end{figure*}

\begin{figure*}[t!]
     \centering
     \begin{subfigure}[b]{0.32\textwidth}
         \centering
         \includegraphics[trim={0 .7cm 0 .7cm},clip,width=\textwidth]{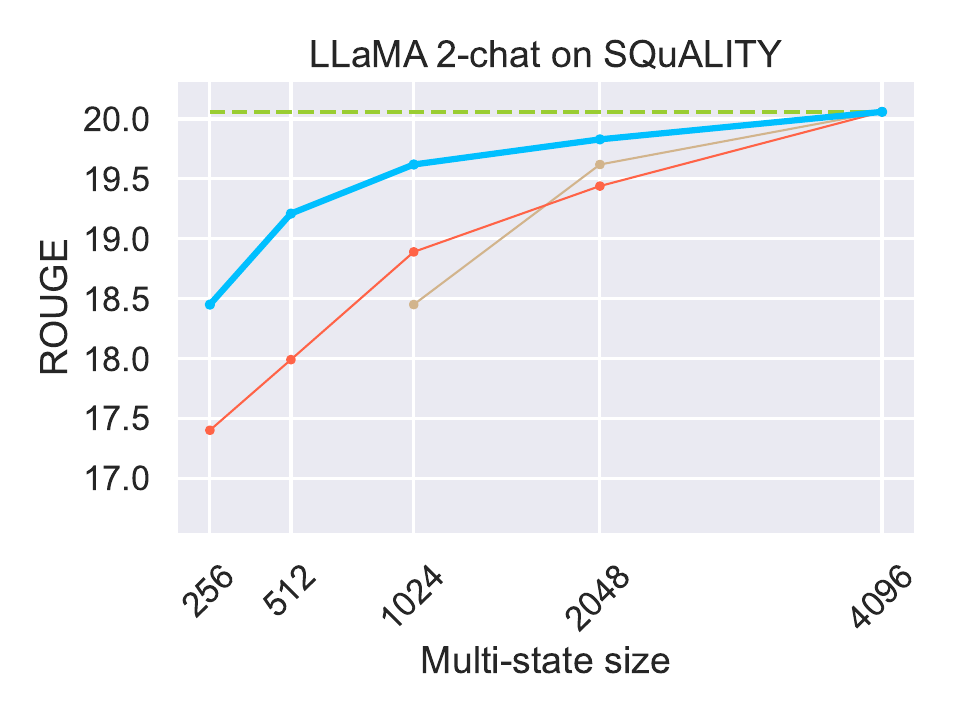}
     \end{subfigure}
     \hfill
     \begin{subfigure}[b]{0.32\textwidth}
         \centering
         \includegraphics[trim={0 .7cm 0 .7cm},clip,width=\textwidth]{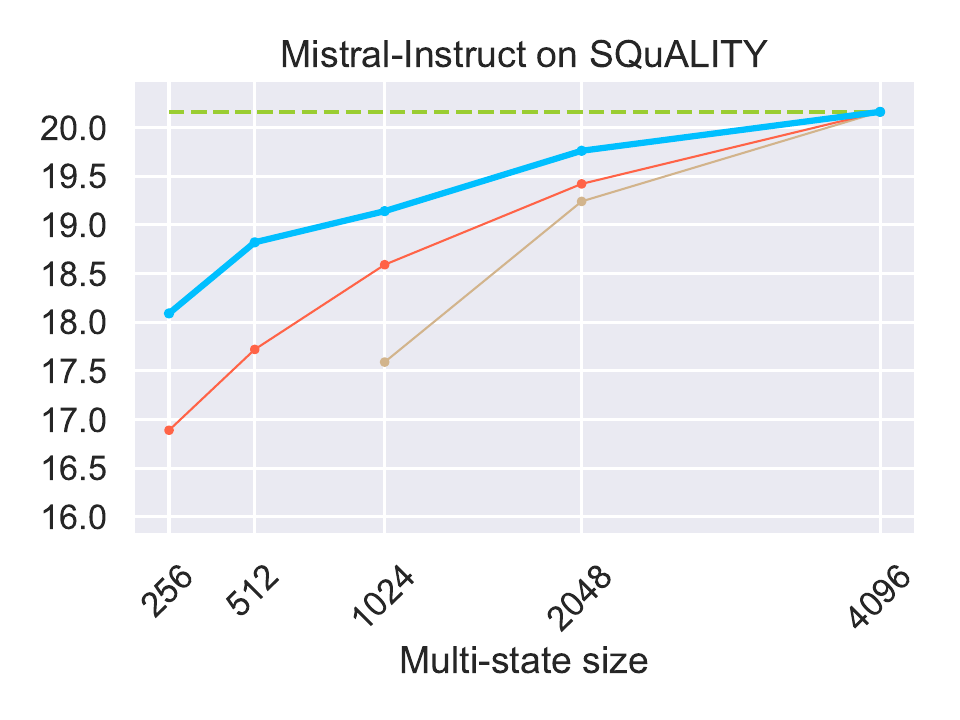}
     \end{subfigure}
     \hfill
     \begin{subfigure}[b]{0.32\textwidth}
         \centering
         \includegraphics[trim={0 .7cm 0 .7cm},clip,width=\textwidth]{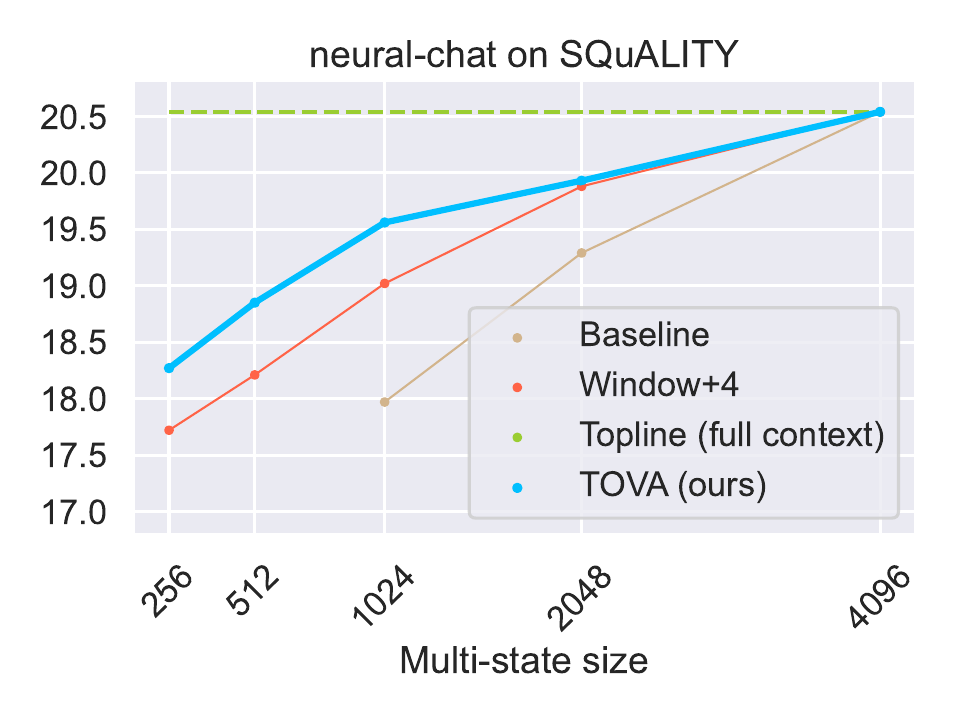}
     \end{subfigure}
     \caption{Geometric mean of ROUGE-1/2/L for SQuALITY. \method achieves within one point of the topline using \(\nicefrac{1}{8}-\nicefrac{1}{4}\) of the multi-state size, while outperforming all other policies.
     \label{fig:squality_ft}}
\end{figure*}

\begin{figure*}[t!]
     \centering
     \begin{subfigure}[b]{0.32\textwidth}
         \centering
         \includegraphics[trim={0 .7cm 0 .7cm},clip,width=\textwidth]{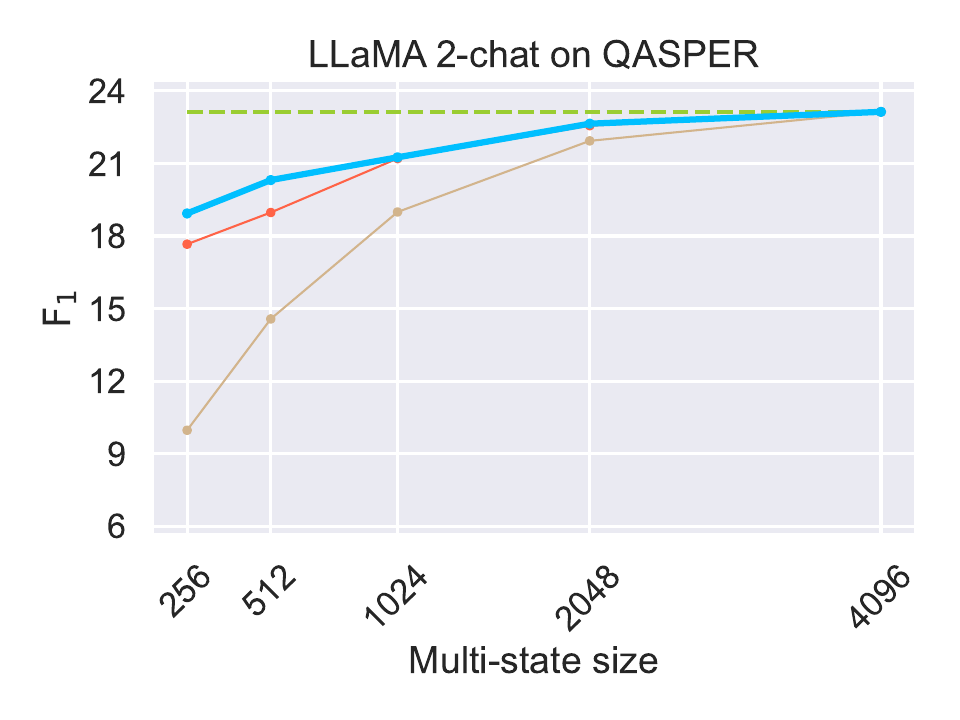}
     \end{subfigure}
     \hfill
     \begin{subfigure}[b]{0.32\textwidth}
         \centering
         \includegraphics[trim={0 .7cm 0 .7cm},clip,width=\textwidth]{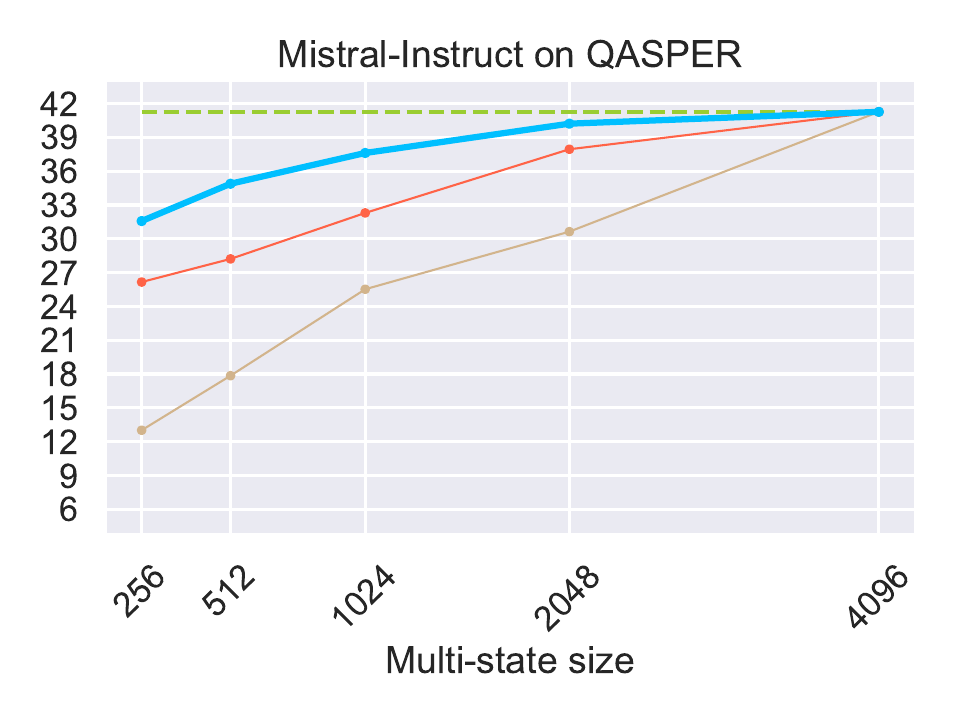}
     \end{subfigure}
     \hfill
     \begin{subfigure}[b]{0.32\textwidth}
         \centering
         \includegraphics[trim={0 .7cm 0 .7cm},clip,width=\textwidth]{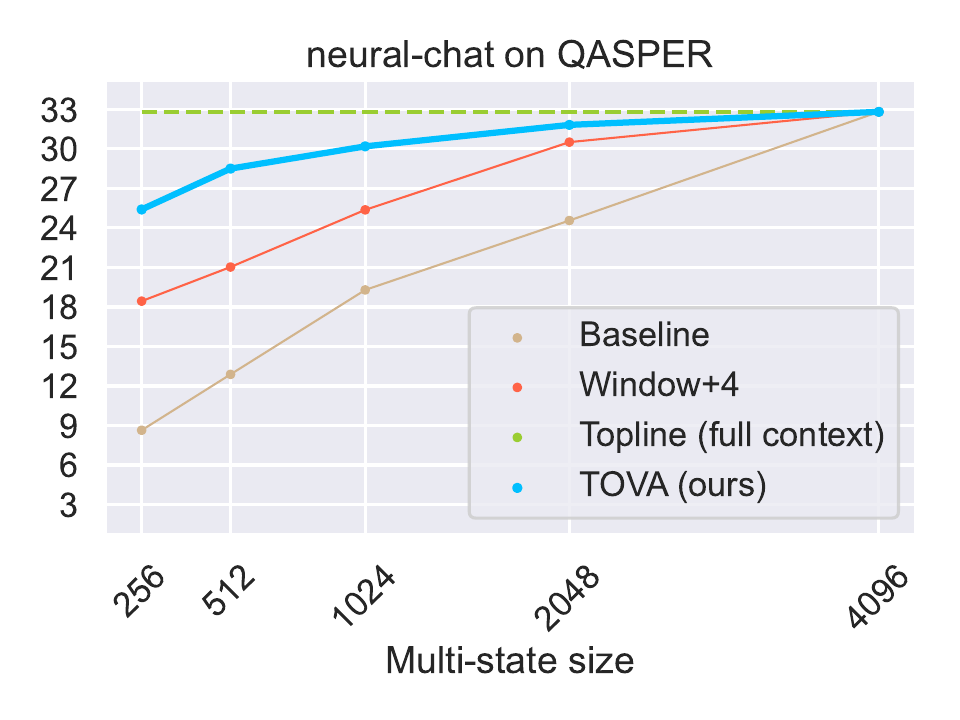}
     \end{subfigure}
     \caption{F1 score over QASPER benchmark. \method outperforms both baselines, but requires a half of the full multi-state size for obtaining comparable results to the topline.
     \label{fig:qasper_ft}}
\end{figure*}

\newparagraph{Text generation} We prompt the models 
to generate a long story. We sample 100 unique stories from each version of the model, using different seeds. As comparing between stories is hard, we employ GPT-4 as an evaluator~\cite{vicuna2023,zhou2023lima}. For each seed, we compare the two generated stories by asking GPT-4 which is better, reporting the average win rate for each approach. For further implementation details, see \cref{sec:generation_app}.

\subsection{Models}
\label{subsec:models}
For language modeling, we experiment with three leading transformer decoder LLMs families, each offering a \(\sim\)7B parameter version: LLaMA-2~\cite{touvron2023llama2}, Mistral~\cite{jiang2023mistral} and Yi~\cite{yi}. %
For long range understanding tasks, we consider three fine-tuned LLMs, which have been shown to excel in instruction tasks: LLaMA-2-chat~\cite{touvron2023llama2}, Mistral-Instruct~\cite{jiang2023mistral} and neural-chat~\cite{lv2023neuralchat}. For text generation, we use MythoLogic~\cite{padar2023mythologic}, a LLaMA-2-13B version fine-tuned for story generation.

For all models and tasks, we use the full training sequence length of 4,096 tokens.
For language modeling, we split the texts into chunks of length 4,096, and apply efficient masking (see \cref{sec:lm_exp_det}). For the language understanding tasks, we truncate the end of the example (excluding prompt) if it exceeds 4,096 tokens, as in~\citet{zeroscrolls}.

%% file: 5_results.tex
\subsection{Language Modeling}
\label{subsec:language_modeling}
We evaluate our base models over the language modeling task using the following policies: \window, \windowi{4}, \hto and our \method policy.\footnote{We ablate other policies in~\cref{sec:abalation_app}.} As an additional baseline, we run the models with a smaller sequence length, while not applying compression, which corresponds to an \infinite \msrnn with a shorter sequence length.
We examine multi-state sizes in exponential scales of 2\(^j\) for \(j \in \{ 6, 7, \dots, 12 \} \)~(2\(^{12}\)=4,096). 

\Cref{fig:ppl} shows the perplexity results on PG-19. 
In all cases, \method performs within 0.4 points of the topline using one eighth of the full context length.
Our results are consistently better than all baselines, which require at least half of the full context length to reach the full model results. 
Based on our results, we consider two policies for the other tasks: \method and \windowi{4}, our best baseline.

\subsection{Long Range Understanding}
\label{subsec:scrolls}
We evaluate instruction-tuned LLMs on SQuALITY and QASPER.\footnote{Base LLMs numbers are reported in \cref{sec:leu_app}.} As an additional baseline, we present the model with a truncated version of the example according to the MSRNN capacity. E.g., for a multi-state of size $k$, the example is truncated to $k$ tokens (including the prompt). As multi-state sizes, we consider 2\(^j\) for \(j \in \{ 8, 9, \dots, 12 \} \).

Results for SQuALITY are shown in \cref{fig:squality_ft}. \method consistently outperforms all baselines across all setups. As in language modeling, \method requires a quarter~(Mistral and Yi) or even one eighth~(LLaMA-2) of the full context to reach within one point of the topline.

\Cref{fig:qasper_ft} shows the QASPER results. The gap between \method and the baselines is large, in some cases reaching beyond 5~F1~points. Nonetheless, here \method needs half of the full context to perform within one F1 point of the topline, and outperforms all baselines across all multi-state sizes.

\subsection{Text Generation}
\label{subsec:text_gen}

\begin{figure}[t!]
     \centering
     \includegraphics[clip,trim={1.25cm 1.25cm .75cm 1.25cm},width=0.91\columnwidth]{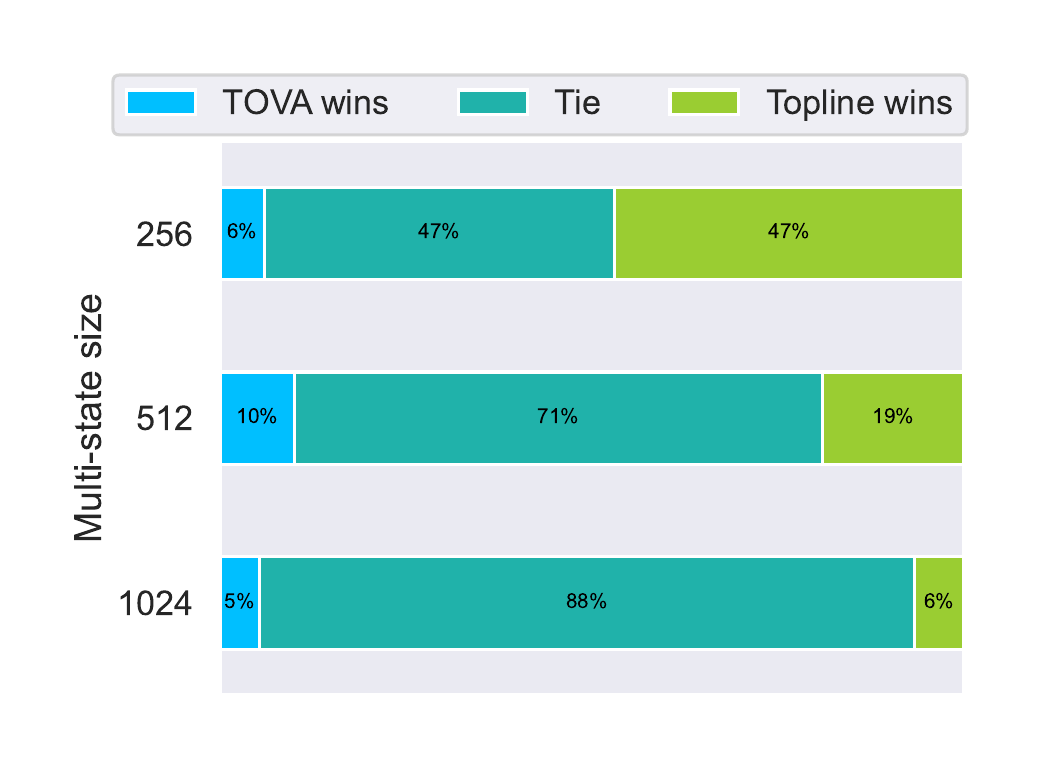}
     \caption{GPT-4 preference over stories generated by the full model and using \method.}
     \label{fig:text_gen}
\end{figure}

We compare \method to the topline on text generation.
We first note that limiting the multi-state size makes the generated text shorter: the average story length for the full model is 1,566 tokens. This value is kept for a multi-state size of 1,024,  but drops to 1,503 with 512 tokens and to 1,361 with 256 tokens. 

\Cref{fig:text_gen} shows the evaluation results of the stories using GPT-4. Using 256 tokens our policy losses to the topline in 47\% of cases, while winning or tying in the remaining cases. This loss rate decreases substantially to 19\% with 512 tokens and further to only 6\% with 1,024 tokens. Importantly, our policy is also preferred over the topline in 5--10\% of the cases in all multi-state sizes considered. 

\subsection{Discussion}
Our results indicate that transformer decoder LLMs often behave empirically as \finite \msrnn: in 2/4 tasks, using \method with as little as \nicefrac{1}{8}--\nicefrac{1}{4} of the multi-state size yields comparable results to the topline. The other two tasks, text generation and retrieval QA, seem to require larger multi-state sizes, though still maintain comparable performance using half of the full multi-state. 
This suggests that the conversion of a transformer into an RNN reintroduces the inherent challenges associated with RNNs, as they encounter difficulties with retrieving distant information~\cite{hochreiter1997long, arjovsky2016unitary, jelassi2024repeat}.

%% file: 6_analysis.tex
We analyze \method in terms of memory and throughput efficiency (\cref{subsec:memory}), extrapolation (\cref{subsec:extra}), and the tokens frequently kept by it~(\cref{subsec:method_ana}). Throughout the section we use LLaMA-2-7B.

\subsection{\method is Time- and Memory-Efficient}
\label{subsec:memory}

\begin{table}
\small
\begin{center}
\begin{tabularx}{\columnwidth}{lccccc}
\toprule
Multi- & 256 & 512 & 1,024 & 2,048 & 4,096 \\
state size & & & & & (full) \\
\midrule
Memory (Gig.) & 0.15 & 0.28 & 0.56 & 1.11 & 2.18 \\
\midrule
\midrule
Maximal batch & \phantom{.}139 & \phantom{.0}70 & \phantom{.0}35 & \phantom{.0}17 & \phantom{.00}8 \\
Rel.~throughput &  \phantom{0}8.5 & \phantom{0}4.8 & \phantom{0}3.1 & \phantom{0}1.7 & \phantom{00.}1 \\
\bottomrule
\end{tabularx}
\end{center}
\caption{\method substantially reduces memory requirements~(first row), and accordingly allows for increased batch size~(second) and throughput~(third row). 
The first row is with a batch size of 1; the second row shows the maximal batch size for decoding the same number of tokens on a single V100 machine.
The last row is the overall decoding throughput when the maximum batch size is employed, relative to a full multi-state size. %
\label{tab:eff} 
}
\end{table}

As discussed in~\cref{subsec:background_trans}, caching the \(K,V\) matrices in transformer autoregressive decoding is common in current frameworks. %
When employing \method as a cache compression policy, these matrices are compressed, which
leads to a proportional reduction in memory requirements~(\cref{tab:eff}, first row).
Importantly, beyond the the KV cache, the LLM decoding memory consumption is determined by two additional factors: the model size~(e.g., number of layers, hidden size), and the batch size. As the former is fixed, caching effectively limits the inference batch-size. \Cref{tab:eff} presents the maximum batch size that can be used in our setup for decoding sequences of length 4,096, along with the corresponding throughput (tokens/sec) while decoding 512 sequences (totaling 2M tokens). 
\method with a multi-state of 512, which performs comparably to the full (4,096) model~(\cref{sec:results}),  allows almost a 9X increase in batch size, and a corresponding speedup of 4.8X compared to the full model. 

\subsection{Extrapolation with \method}
\label{subsec:extra}

\begin{figure}[t]
     \centering
     \includegraphics[clip,trim={.25cm .25cm .25cm .25cm},width=.90\columnwidth]{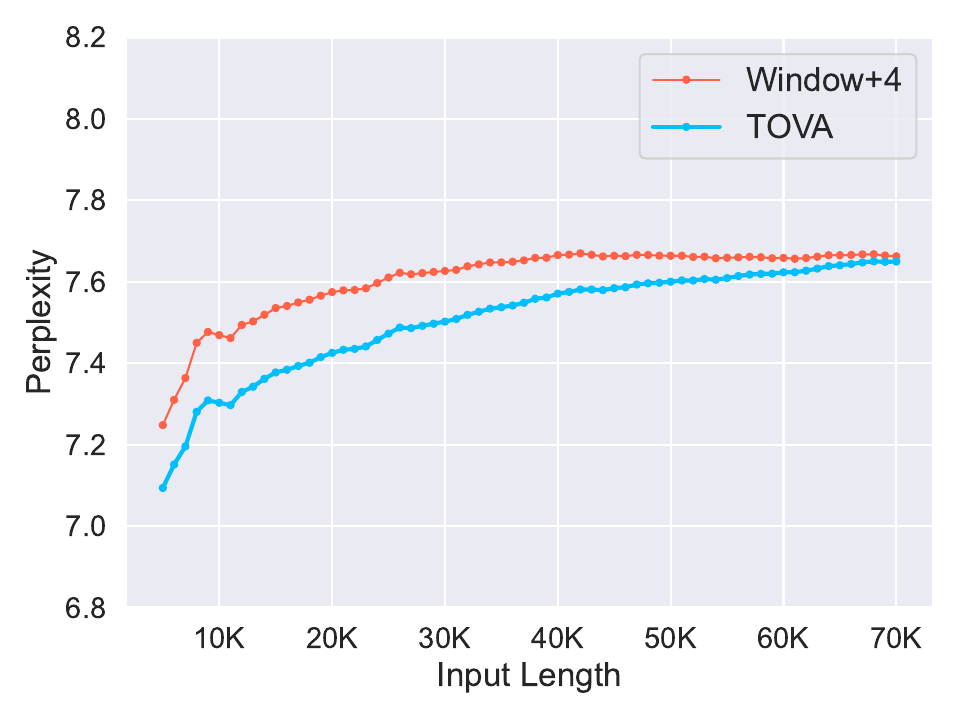}
     \caption{\method successfully extrapolates well beyond pretraining context length, and  outperforms \windowi{4}. Each point is the average over all previous tokens.}
     \label{fig:extrapolation}
\end{figure}

We further test the ability of \finite \msrnns in handling longer texts, i.e., beyond the training sequence length. 
Using \method, this requires adapting the positional encoding of cached tokens to avoid values unseen during training. To do so, we compress the gap $g$ between adjacent token representations to be $ln(ln(g))$,\footnote{Preliminary experiments with other compression functions, e.g., $ln(g)$ and $sqrt(g)$ showed inferior results.} %
while  keeping $g$ fixed if $g\le 10$ to retain local sensitivity. E.g., for adjacent tokens with positions ($i,i+g$), the new positions will be ($i,i+ln(ln(g))$), or ($i,i+g$) if $g \le 10$.

We report the average perplexity on the first 70K tokens of all PG-19 books with at least that number of tokens (52 books in total). We use a multi-state size of 512.  As models struggle to extrapolate to such long contexts, we only compare \method with \windowi{4}, which has been shown to support such contexts~\cite{streamllm, lminfinite}. Our results~(\cref{fig:extrapolation}) show that \method extrapolates well up to 70K tokens with a similar performance to the shorter contexts (less than 0.5 PPL points difference), while outperforming \windowi{4}.

\subsection{Which Tokens Matter?}
\label{subsec:method_ana}

\begin{figure}
     \centering
     \includegraphics[width=.75\columnwidth]{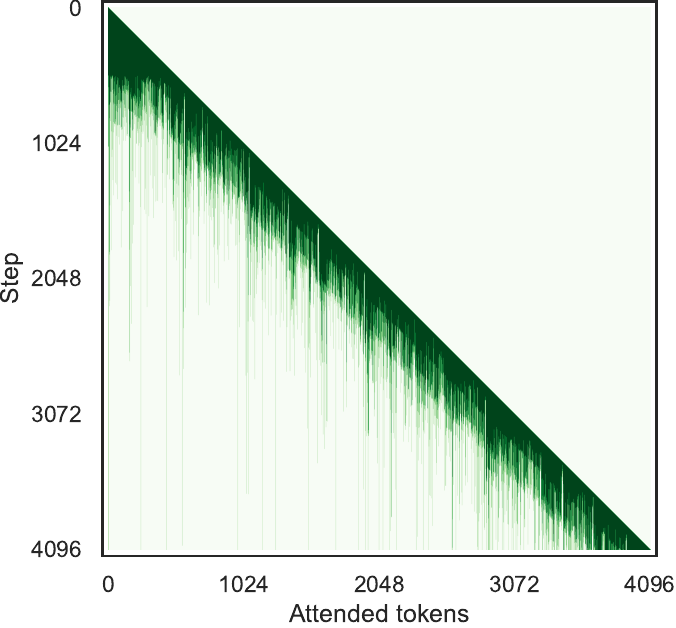}
     \caption{The tokens kept by \method in the final layer of LLaMA-2-7B on one PG-19 example. Rows represent decoding steps, and columns tokens attended to.}
     \label{fig:atten_mask}
\end{figure}

Our results indicate that most tokens may be dropped from memory as generation progresses. We characterize the tokens frequently dropped by running \method on 31 PG-19 instances. 

\newparagraph{Recency is \textit{not} all you need}
Much like most compression policies~(\cref{subsec:baselines})\resolved{\michael{flag to change to 4.1 if we stick with the change}}, \method preserves recent tokens. \Cref{fig:atten_mask} illustrates the tokens kept by \method in the final layer for one PG-19 example, using a multi-state size of 512.\footnote{See \cref{sec:extended_analysis_app} for the illustrations of all layers.}
We see a clear window trend, indicating the importance of recent tokens. Nonetheless, we also observe that many older tokens are kept. To quantify this, we compute the proportion of recent tokens of all tokens kept in the multi-state, averaged across examples, layers, and positions. We find that only 73--76\% of the tokens are recent. This suggests that while recent tokens are important, they are far from sufficient. Importantly, unlike existing policies that handcraft the recent window~\cite{streamllm, hto}, \method 
identifies it automatically.
We turn to study which early tokens tend to be kept, considering two dimensions: position and content.

\newparagraph{The first token matters}

\begin{figure}
     \centering
     \includegraphics[clip,trim={.25cm 0 1.5cm 1.5cm},width=0.91\columnwidth]{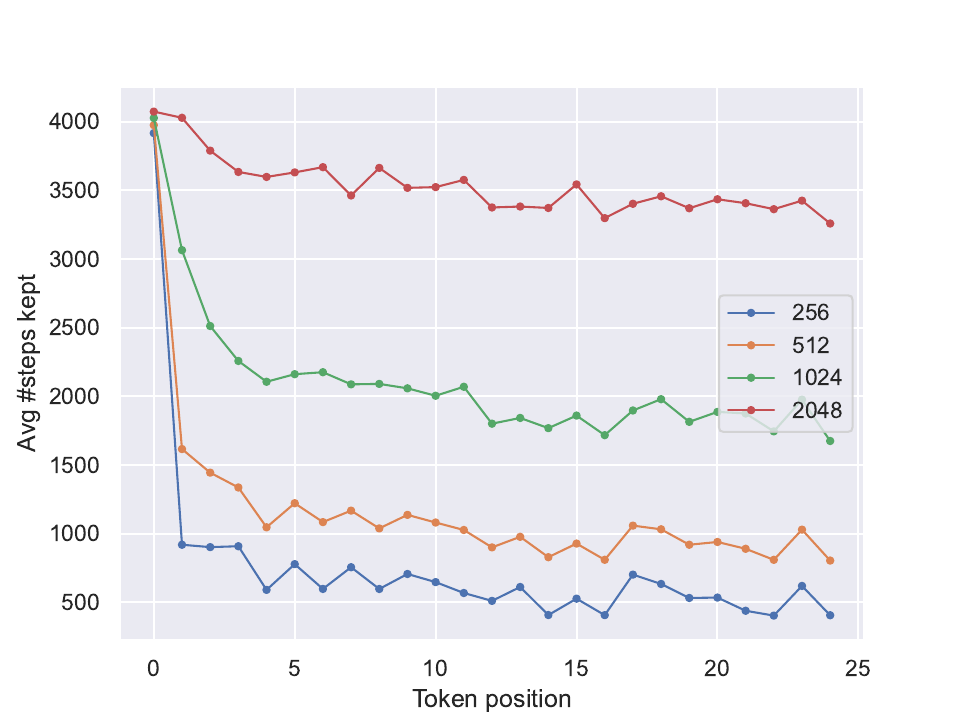}
     \caption{The average number of steps a token is kept in the multi-state when applying \method as a function of token position. Different lines are different multi-state sizes. The very first token is kept through the entire context, while next tokens are dropped far earlier.}
     \label{fig:first_token}
\end{figure}

\Cref{fig:first_token} shows the number of decoding steps each of the first 25 tokens is kept (averaged across layers and examples). 
As previously observed~\cite{lminfinite, streamllm}, we find that the first token is kept until the end of the sequence across all multi-state sizes. 
However, other early tokens are dropped far faster.

\newparagraph{Not all tokens are equally kept}

As indicated by~\cref{fig:atten_mask}, some tokens last much longer than others. To further study it, we map each token to its part-of-speech tag using NLTK~\cite{bird2009natural}, and plot the tags that last longest in~\cref{tab:mean_attention_per_pos_tag}. %
Our results show that, as observed by previous work~\cite{clark-etal-2019-bert,hto,ge2023model}, punctuation and other special symbols tend to be kept. However, we also identify other tokens that tend to stay longer, e.g., possessive endings~(POS) and proper nouns (NNPS). %
Studying the role of these tokens is an exciting direction for future work.

\com{
\michael{alternate 6.1: 
We have shown so far that pretrained transformer LLMs, trained as \infinite \msrnns, in many cases behave empirically as \textit{\finite} \msrnns. This effectively means that most of the tokens are dropped from memory as generation progresses. We turn now to analyze which tokens are kept and which omitted by \method, to understand better its functionality, and which states are important for LLMs to function well.
The analysis below uses an average of 31 instances from PG-19. \\
We first observe that, like most compression policies~(\cref{subsec:baselines}), \metod preserve recent tokens. \cref{fig:atten_mask} illustrates the tokens kept by the \method policy in the last layer for one PG-19 example. The corresponding illustrations for all 32 layer are presented in~\cref{sec:extended_analysis_app}. 
\\
As can be seen, there is clear window trend which indicates that recent tokens are important for the model.  Nevertheless, on average, only 73-76\% are recent tokens, which suggests that while recent tokens are important, they are far from sufficient, as $\approx$\nicefrac{1}{4} of the tokens are chosen based on their content rather than their position. Next we investigate the types of this special tokens.
**need to incorporate somewhere that \method gets this window implicitly while other predetermine it**
\\ \\
Next, we 
}
}

\begin{table}[t!]
\small
\begin{center}
\begin{tabular}{lrrrr}
\toprule
\diagbox[innerwidth=\textwidth*1/7]{Tag}{Multi-state \\ size} & 256 & 512 & 1024 & 2048 \\
\midrule
Avg. & 249 & 481 & 897 & 1537 \\
\midrule
POS & 1134 & 1393 & 1736 & 2061 \\
'' & 845 & 1101 & 1413 & 1774 \\
\$ & 329 & 724 & 1276 & 2123 \\
) & 379 & 670 & 1161 & 1558 \\
. & 350 & 645 & 1117 & 1677 \\
NNPS & 321 & 578 & 1042 & 1671 \\
\verb|\n| & 303 & 550 & 969 & 1538 \\
\bottomrule
\end{tabular}
\end{center}

\caption{\label{tab:mean_attention_per_pos_tag} 
Mean number of steps tokens are kept in the multi-state with \method, grouped by part-of-speech tags. Columns represent the multi-state size.
Here we report the tokens kept the longest, see full table in~\cref{sec:pos_app}.}
\end{table}

%% file: 7_related_work.tex
\resolved{\roy{note that you add a period following paragraph titles in some cases but not all. I generally prefer without it, though adding is fine, as long as you are consistent.}}
\newparagraph{Transformers and RNNs} Several works have tried to bridge the gap between RNNs and transformers. \citet{hutchins2022blockrecurrent} employed a hybrid approach that attends both to recent tokens and to further hidden states. %
\citet{sun2023retentive} substituted the self-attention layer with a convolution layer that can be applied recurrently.
\citet{rwkv} adjusted the self-attention layer to perform recurrence at inference.~\citet{so2022primer} proposed a model with repeated layers to perform recurrent computations over the signal, rather than over time.

Most related to this work are~\citet{katharopoulos2020transformers} and~\citet{peng-etal-2022-abc}.
The former suggested that transformers can be used in a recurrent manner, and proposed a linear transformer for doing so. The latter presented transformers with bounded memory, showing that several transformer variants such as Linformer~\cite{wang2020linformer} and window attention can be interpreted as instances of their framework. Unlike us, these works treat the memory as a single state, without an explicit mapping from tokens to states.
Moreover, unlike our approach, %
the works above require a dedicated training, and cannot operate on existing LLMs.

\newparagraph{Limited KV cache}

Window attention~\cite{wang-etal-2019-multi,beltagy2020longformer,zaheer2020big} and its variants~(e.g., \hto,~\citealp{hto}; SCISSORHANDS,~\citealp{liu2024scissorhands}) are simple ways of limiting the cache size in transformers. A recent followup work \cite{ge2023model} showed that manually caching specific tokens like ``.'' and ``,'' further boosts \hto performance. We showed that \method does so without manually selecting tokens~(\cref{subsec:method_ana}). \citet{anagnostidis2023dynamic} introduced a learned approach over LLMs that limits the cache consumption of transformers. \citet{yun-etal-2023-focus} and \citet{berchansky2023optimizing} proposed token pruning and token combining.

Concurrent to our work, \citet{ren2024efficacy} suggested robustness measures to choose which states to drop; \citet{brandon2024reducing} showed that KV cache can be shared across layers;  \citet{yang2024pyramidinfer} proposed a pyramid structure across layers to reduce cache size; \citet{li2024snapkv} and \citet{zandieh2024subgen} suggested clustering the KV cache, and  \citet{kang2024gear} proposed to quantize and approximate it. None of these works drew a connection between RNNs and transformers.

\newparagraph{New RNN variants} 
Recent work aimed to revive RNNs in NLP. S4~\cite{gu2022efficiently} and its successors~\cite{dss,gss,mamba} elevate state spaces to form linear RNNs. Other work introduced RNN variants that train effectively~\cite{merity2019single, orvieto2023resurrecting, yang2023gated,beck2024xlstm}.
\resolved{\roy{another relevant citation (and fun paper to read) https://arxiv.org/abs/1911.11423}}

\newparagraph{Simplifying transformers}
Previous work has shown that many transformer attention heads can be pruned~\cite{Michel2019sixteen,li-etal-2021-differentiable} or replaced with static weights~\citep{hassid-etal-2022-much}. Several works replaced the attention mechanism in transformers with efficient variants~\cite{peng2021rfa,performer,liu-2021-gmlp, lee-thorp-etal-2022-fnet}. We show that transformer decoders can be reduced to \finite \msrnns.

%% file: acknowledgements.tex
We thank Miri Varshavsky Hassid for the great feedback and moral support. This work was supported in part by NSF-BSF grant 2020793.

%% file: app_ablation.tex
\begin{table*}[th!]
\centering
\begin{tabular}{c|ccccccc}
\diagbox[innerwidth=\textwidth*1/7]{Policy}{Multi-state \\ size} & 64 & 128 & 256 & 512 & 1024 & 2048 & 4096 \\
\midrule
Baseline  & 17.65 & 12.97 & 10.39 & 8.92 & 8.04 & 7.50 & \textbf{7.16} \\
\window   & 4812.27 & 4025.01 & 3275.58 & 2184.62 & 1001.29 & 240.17 & \textbf{7.16} \\
Window\(+1\)   & 10.20 & 8.97 & 8.22 & 7.76 & 7.50 & 7.33 & \textbf{7.16} \\
Window\(+4\)   & 10.28 & 8.98 & 8.19 & 7.73 & 7.46 & 7.30 & \textbf{7.16} \\
\hto-head  & 10.22 & 8.97 & 8.21 & 7.75 & 7.49 & 7.32 & \textbf{7.16} \\
\hto-layer   & 10.20 & 8.97 & 8.22 & 7.76 & 7.50 & 7.33 & \textbf{7.16} \\
\method-head   & 11.13 & 9.55 & 8.69 & 7.90 & 7.52 & 7.27 & \textbf{7.16} \\
\method-layer   & \textbf{9.53} & \textbf{8.32} & \textbf{7.71} & \textbf{7.41} & \textbf{7.25} & \textbf{7.17} & \textbf{7.16} \\
\method-layer\(+1\)   & \textbf{9.53} & \textbf{8.31} & \textbf{7.71} & \textbf{7.41} & \textbf{7.25} & \textbf{7.17} & \textbf{7.16} \\
\method-layer\(+4\)  & 9.63 & \textbf{8.33} & \textbf{7.72} & \textbf{7.41} & \textbf{7.25} & \textbf{7.17} & \textbf{7.16} \\
\end{tabular}
\caption{Perplexity over the PG-19 set using varying multi-state sizes (maximal number of states used), while ablating several dimensions such as the number of recent tokens in \windowi{i} policies and head vs.~layer selection in \hto and \method. Our \method policy dominates the table in all multi-state sizes.\label{tab:lm_ablation}} 
\end{table*}

We ablate all policies presented in \cref{subsec:baselines} and several \method variants with the language modeling task. Specifically we examine: \window, \windowi i for \(i \in \{1,4\}\), \hto for both per layer and per head approaches and our \method policy for both per layer and per head approaches. We also combine \method with additionally fixing the first \(i\) tokens using \(i \in \{1,4\}\). We consider the same baseline policy as in \cref{subsec:language_modeling}. We use the LLaMA-2-7B as the backbone model.

Our results are presented in \cref{tab:lm_ablation}. As shown in \cref{subsec:language_modeling} the \window policy fails, while the \windowi{1} and \windowi{4} policies maintain much better results (with \windowi{4} performing slightly better). The two \hto policies (head/layer) produce similar results. Regarding our \method policies, the head version performs worse than former policies in most multi-state sizes, while the layer version outperforms all other policies. We attribute this difference to the more robust selection mechanism employed by the layer version, which requires agreement among all heads to determine the importance of specific tokens. Lastly, when we enhance our \method policy with the explicit preservation of \(i\) initial tokens, the results remain relatively unchanged, implying that our policy inherently retains the crucial tokens.

%% file: app_algo.tex
\section{Formal Description of Method}
\label{sec:tova_algo}

\Cref{algorithm:alg_1} provides a torch-like implementation of the \method cache compression policy.

\begin{algorithm*}
\footnotesize{
\vspace{+0.1cm}
\begin{minted}{python}
def TOVA(attn_weights, k_cache, v_cache, cache_max_size): 
    # k_cache.shape and v_cache.shape are [attn_heads, num_kv, hidden_dim]
    attn_heads, num_q, num_kv = attn_weights.shape
    if num_kv <= cache_max_size:
        return
    # Average last query attention weights across heads:
    mean_attn_weights = mean(attn_weights[:,-1,:], dim=0) 
    minimal_idx = argmin(mean_attn_weights) # get the index to drop
    k_cache = concat([k_cache[:, :minimal_idx], k_cache[:, minimal_idx+1:]], dim=1)
    v_cache = concat([v_cache[:, :minimal_idx], v_cache[:, minimal_idx+1:]], dim=1)
\end{minted}
\caption{A torch-like implementation of \method. Batch size=1 is assumed for simplicity.\label{algorithm:alg_1} %
}
}
\end{algorithm*}

%% file: app_prompts.tex
The prompts used for the different evaluations through this work are presented in \cref{tab:prompts}.

\begin{table*}[t]
    \centering
    \begin{tabular}{@{}p{0.165\textwidth}p{0.835\textwidth}@{}}
    \toprule
    \textbf{Task} & \textbf{Prompt}\\
    \midrule
     SQuALITY &  \{Story\}
     
     ~
     
     Answer the question in a paragraph.
     
     ~
     
     Question:
         
    ~
    
    \{Question\}
    
    ~
    
    Answer:   
     \\
    \midrule
     QASPER &  \{Article\}
     
     ~
     
     Answer the question as concisely as you can, using a single phrase or sentence if possible. If the question cannot be answered based on the information in the article, write ``unanswerable''. If the question is a yes/no question, answer ``yes'', ``no'', or ``unanswerable''.
     
     ~
     
     Question:
         
    ~
    
    \{Question\}
    
    ~
    
    Answer:   
     \\

    \midrule
     Story Generation & \#\#\# Instruction:
     ~
     
     Write a very long story (at least 4,000 words). The story should include at least 20 named characters, spans 3 countries and 9 cities, at least 10 chapters and should have a lot of plot twists.    
    
    ~
    
    \#\#\# Response:
     \\

     \midrule
     GPT- Evaluation & Help me decide which response is better given the prompt:
      ~
      
     \{Prompt body for story generation\}
      ~
      
      Which of the following responses is better (the responses are separated by '------------------------'):
     
       ~
     
       Response (A):
            ~
            
     \{First Response\}
     
          ~
     
    ------------------------
    
     ~
     
    Response (B):
     ~
     
     \{Second Response\}

          ~
     
Comparing these two responses, which response is better (A), (B) or (C) for equal quality? please select one and only one option, be as concisely as you can, using a single phrase.
     \\
    \bottomrule
    \end{tabular}
    \caption{
    Prompts used for our experiments.}
    \label{tab:prompts}
\end{table*}

%% file: generation_app.tex
To evaluate the generated texts, using GPT-4, we use the gpt-4-0613 version. We drop cases where the model stops generating before reaching the memory limit, as both stories are identical. To account for GPT-4's positional bias~\cite{wang2023large}, we present each pair of stories twice, alternating their positions, and only consider a win if the same approach is preferred in both cases.

%% file: app_lm_implementation.tex
All experiments are done using bfloat16 floating-point precision over Nvidia V100 GPUs. 
To effectively parallelize the language modeling task for all tokens in the sequence, we modify the attention mask to incorporate the different \msrnn policies presented in \cref{sec:method}. Specifically, for \window and \windowi{i} policies, we apply a static masking, as the reduced tokens are independent with respect to the attention computation. For \hto and \method, we adjust the mask according to the attention weights of the relevant layer.

%% file: app_long_range.tex
\begin{figure*}[t!]
     \centering
     \begin{subfigure}[b]{0.32\textwidth}
         \centering
         \includegraphics[width=\textwidth]{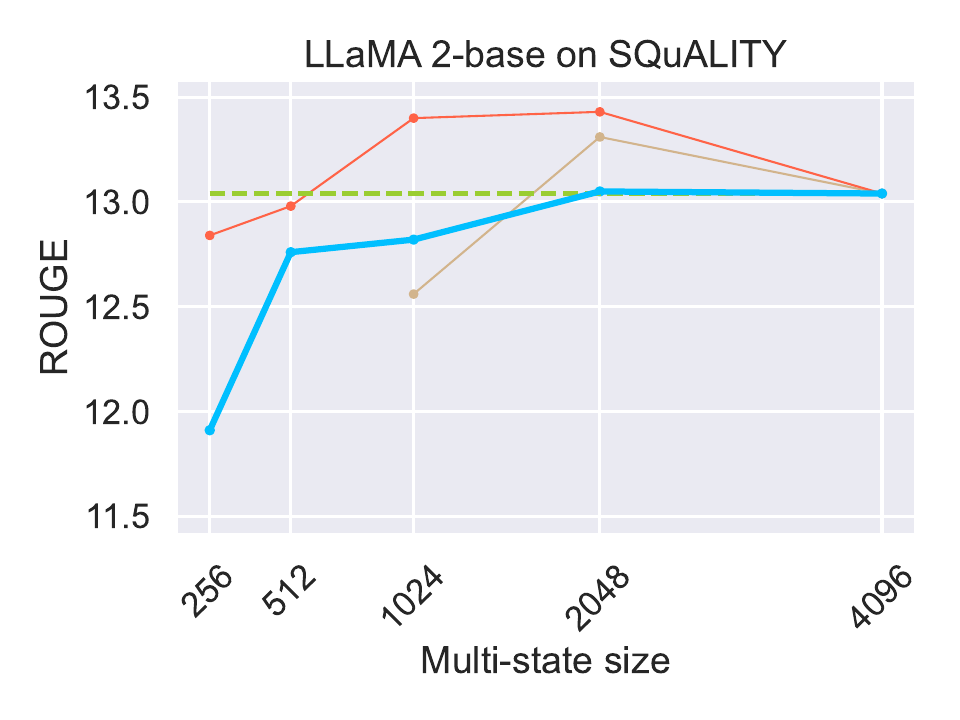}
     \end{subfigure}
     \hfill
     \begin{subfigure}[b]{0.32\textwidth}
         \centering
         \includegraphics[width=\textwidth]{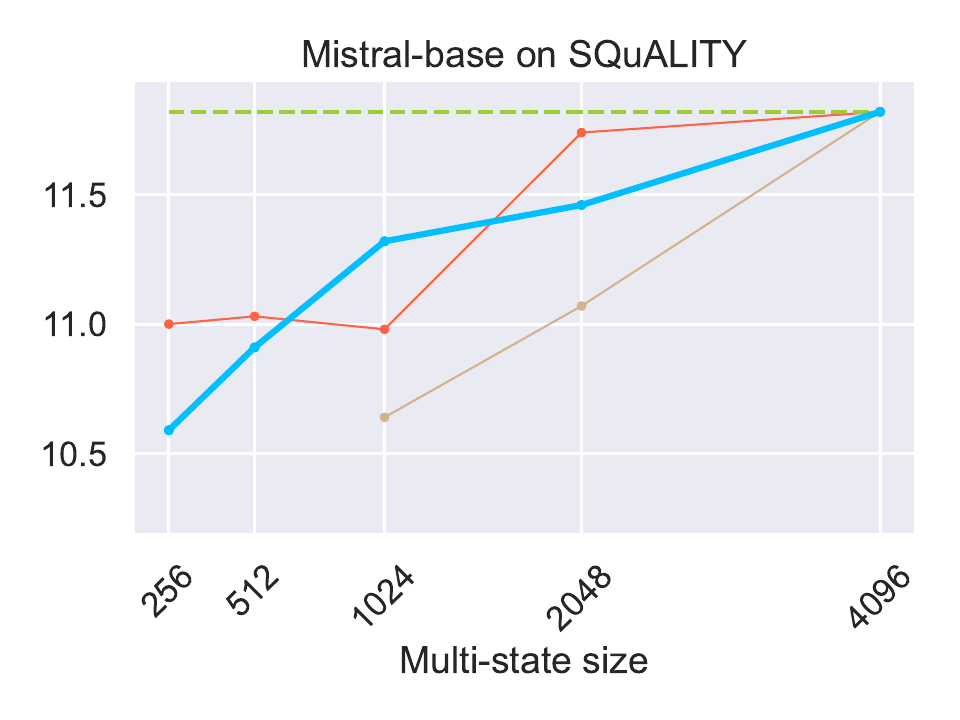}
     \end{subfigure}
     \hfill
     \begin{subfigure}[b]{0.32\textwidth}
         \centering
         \includegraphics[width=\textwidth]{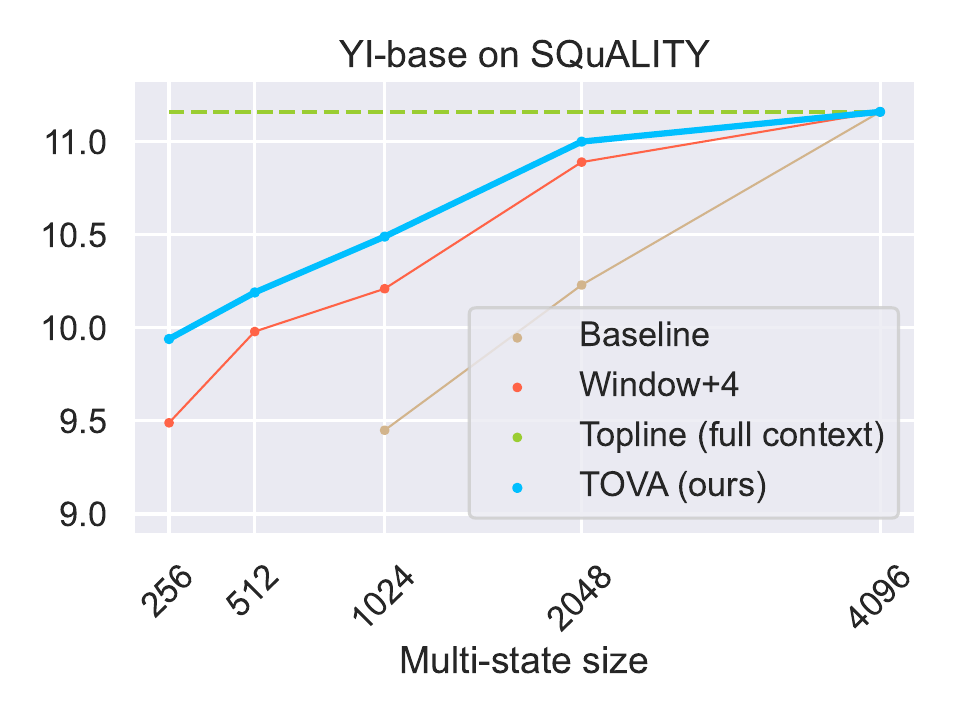}
     \end{subfigure}
     \caption{Geometric mean of ROUGE-1/2/L for SQuALITY using the base LLMs. \label{fig:squality_base}}
\end{figure*}
\begin{figure*}[t!]
     \centering
     \begin{subfigure}[b]{0.32\textwidth}
         \centering
         \includegraphics[width=\textwidth]{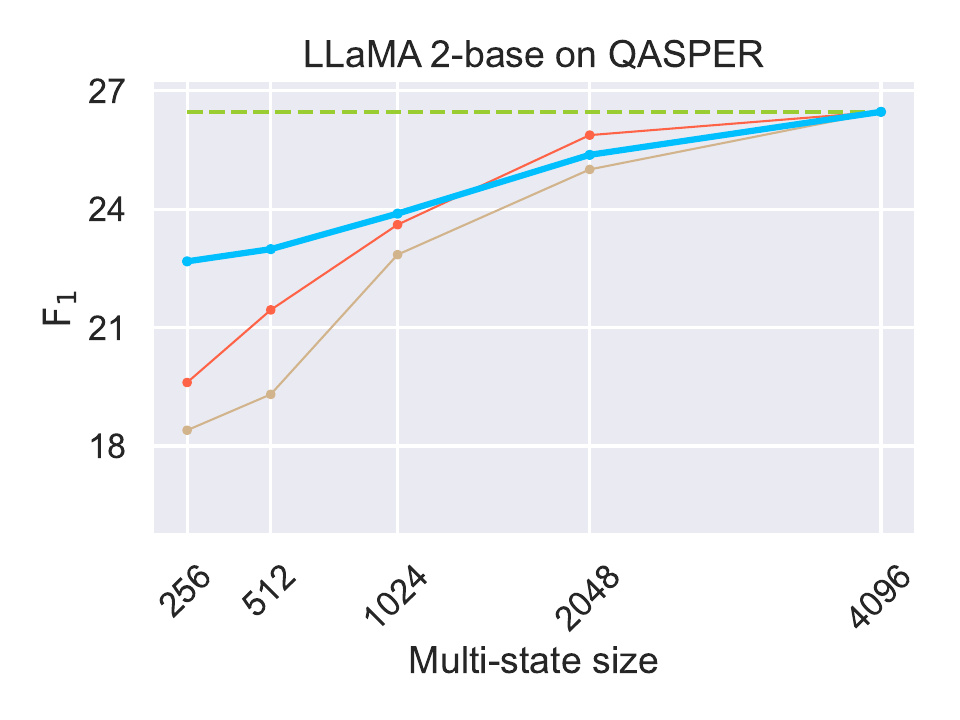}
     \end{subfigure}
     \hfill
     \begin{subfigure}[b]{0.32\textwidth}
         \centering
         \includegraphics[width=\textwidth]{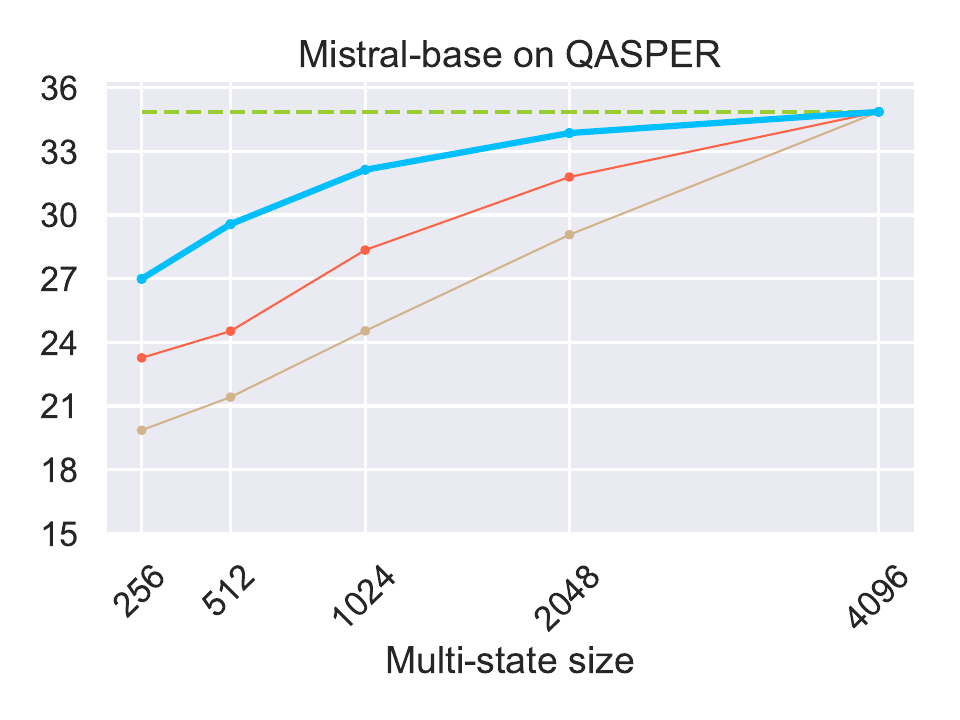}
     \end{subfigure}
     \hfill
     \begin{subfigure}[b]{0.32\textwidth}
         \centering
         \includegraphics[width=\textwidth]{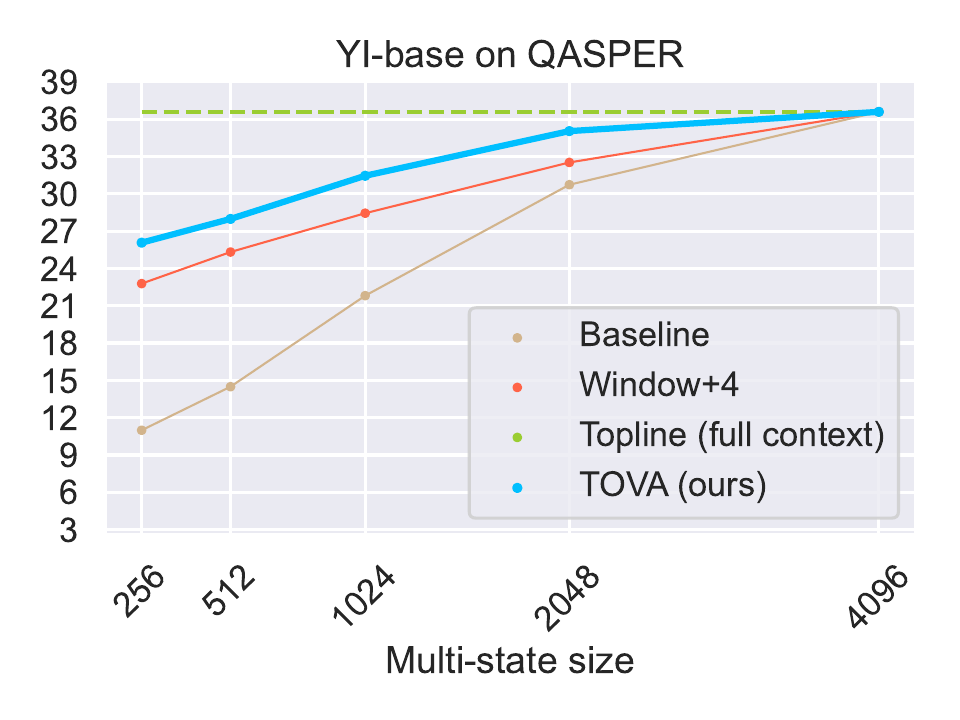}
     \end{subfigure}
     \caption{F1 scores over the QASPER benchmark using base LLMs. \label{fig:qasper_base}}
\end{figure*}

\Cref{fig:squality_base,fig:qasper_base} show the results for base LLMs over the SQuALITY and QASPER benchmarks, respectively.

%% file: app_ext_analysis.tex
\begin{figure*}[bt]
\centering
 \makebox[\textwidth]
{\includegraphics[width=0.87\paperwidth]{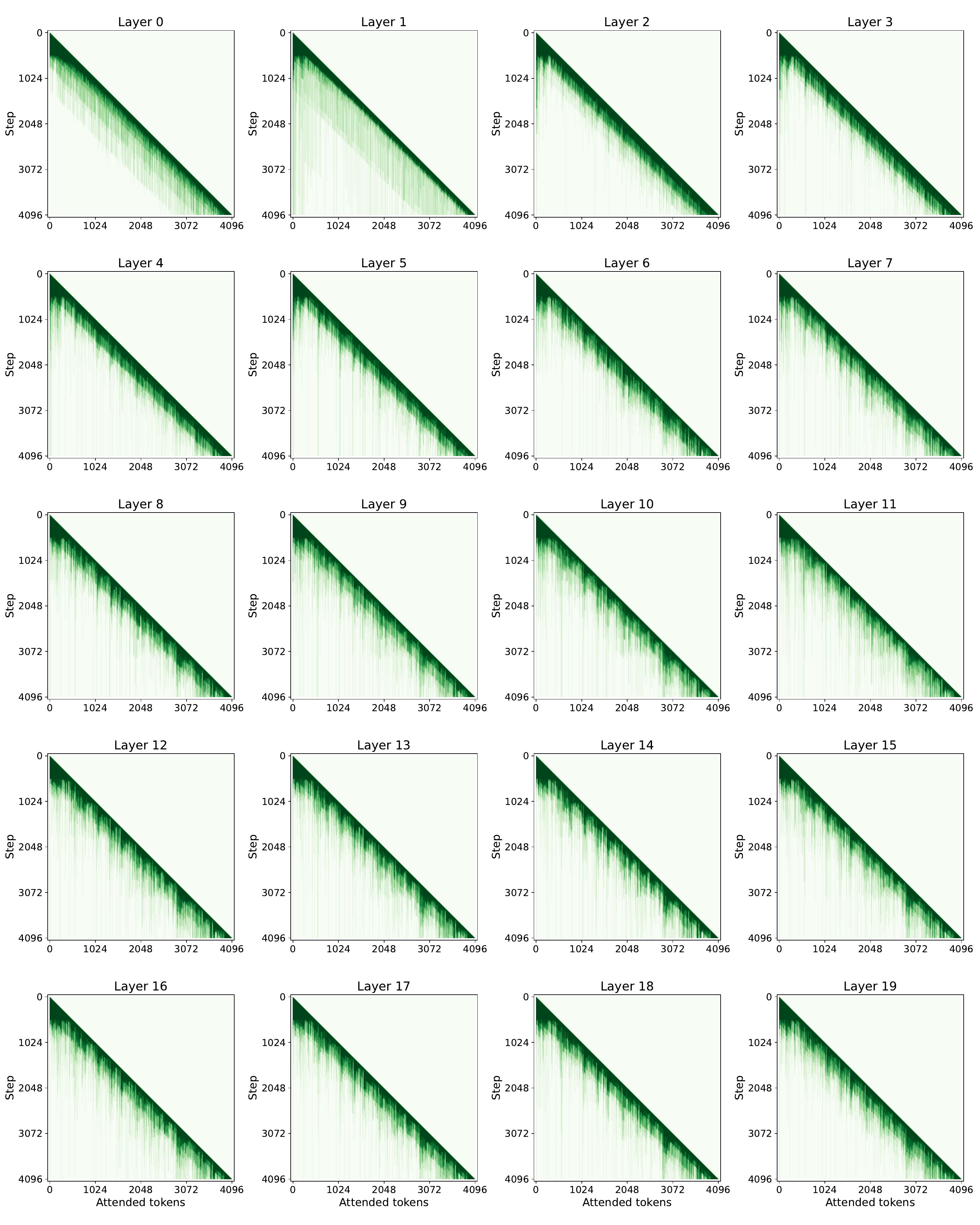}}
\caption{The full illustration corresponding to \cref{fig:atten_mask} of the tokens kept by \method for all layers of LLaMA-2-7B on one PG-19 example. Each row represents a decoding step, and each column is a token attended to. Layers 0--19.
\label{fig:all_layers_1}}
\end{figure*}

\begin{figure*}[t!]
\centering
\vspace{-270pt}
 \makebox[\textwidth]
{\includegraphics[width=0.87\paperwidth]{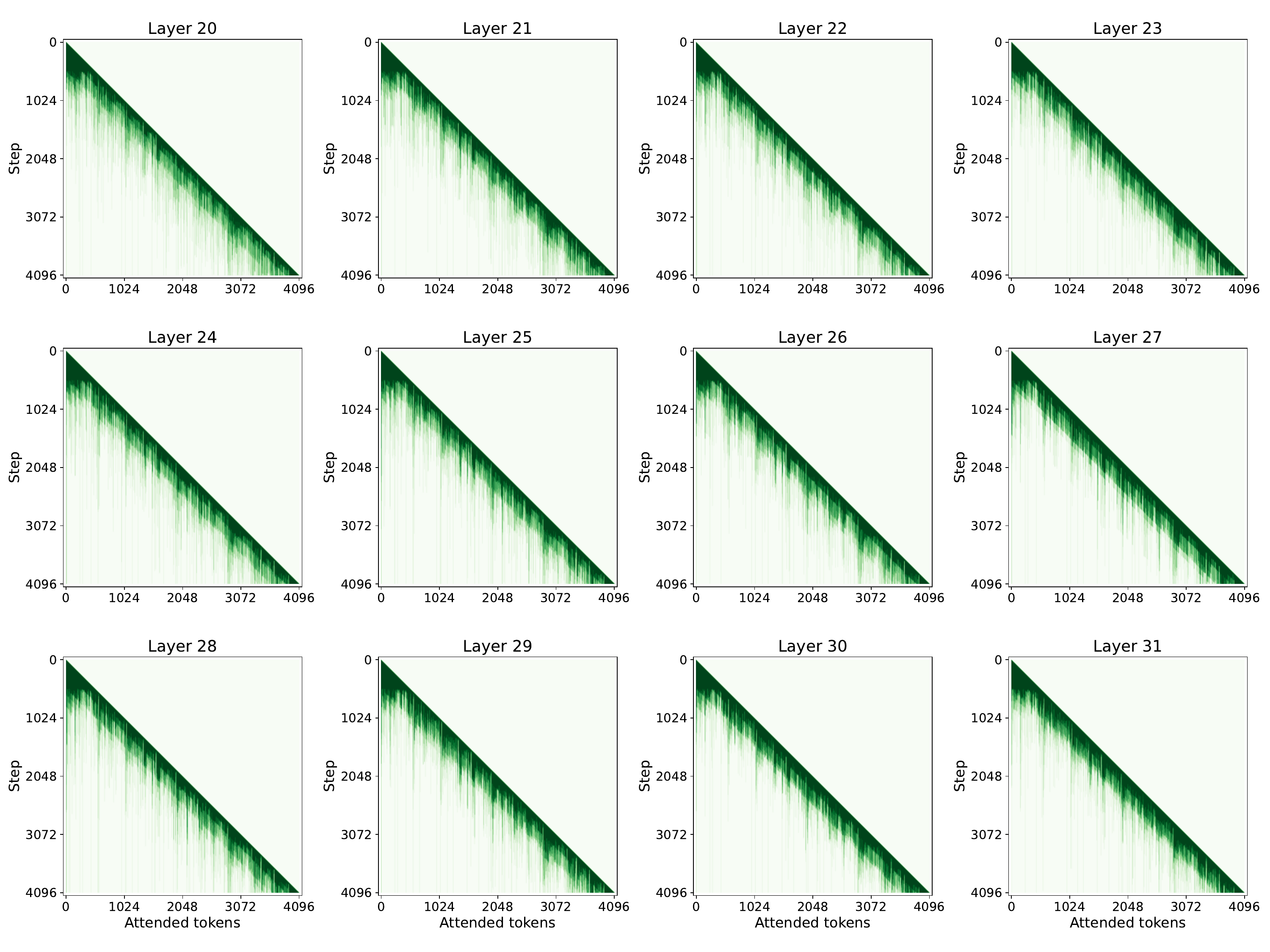}}
\caption{Continuation of \cref{fig:all_layers_1} for layers 20--31. \label{fig:all_layers_2}}
\end{figure*}

\Cref{fig:all_layers_1,fig:all_layers_2} show illustrations of the tokens retained (X axis) at each step (Y axis) for every layer of LLaMA-2-7B, when applying \method over one PG-19 example. We use a multi-state size of 512.

%% file: app_pos.tex
The full version of \cref{tab:mean_attention_per_pos_tag} is presented in \cref{tab:full_pos}.

\begin{table}
\begin{tabular}{lrrrr}
\toprule
\diagbox[innerwidth=\textwidth*1/7]{Tag}{Multi-state \\ size} & 256 & 512 & 1024 & 2048 \\
\midrule
Avg. & 249 & 481 & 897 & 1537 \\
\midrule
POS & 1134 & 1393 & 1736 & 2061 \\
'' & 845 & 1101 & 1413 & 1774 \\
\$ & 329 & 724 & 1276 & 2123 \\
) & 379 & 670 & 1161 & 1558 \\
. & 350 & 645 & 1117 & 1677 \\
NNPS & 321 & 578 & 1042 & 1671 \\
\verb|\n| & 303 & 550 & 969 & 1538 \\
WP\$ & 255 & 539 & 1121 & 1920 \\
CD & 301 & 537 & 940 & 1557 \\
NN & 270 & 527 & 983 & 1628 \\
NNS & 270 & 526 & 978 & 1618 \\
NNP & 270 & 517 & 951 & 1613 \\
FW & 253 & 511 & 903 & 1444 \\
: & 243 & 492 & 940 & 1570 \\
JJ & 240 & 480 & 918 & 1598 \\
VBP & 244 & 478 & 882 & 1504 \\
JJS & 220 & 475 & 953 & 1689 \\
UH & 233 & 474 & 870 & 1412 \\
SYM & 231 & 471 & 893 & 1482 \\
WDT & 223 & 462 & 903 & 1604 \\
VBN & 230 & 462 & 887 & 1549 \\
EX & 244 & 461 & 847 & 1461 \\
RB & 223 & 459 & 892 & 1566 \\
, & 236 & 453 & 840 & 1454 \\
VBG & 221 & 445 & 858 & 1523 \\
RBS & 210 & 441 & 878 & 1645 \\
VBZ & 219 & 440 & 844 & 1492 \\
CC & 217 & 437 & 862 & 1546 \\
VBD & 217 & 432 & 827 & 1493 \\
VB & 214 & 426 & 817 & 1457 \\
PRP & 217 & 424 & 794 & 1432 \\
RP & 207 & 417 & 811 & 1485 \\
WRB & 207 & 415 & 800 & 1502 \\
WP & 199 & 405 & 803 & 1506 \\
JJR & 195 & 403 & 782 & 1413 \\
RBR & 183 & 397 & 821 & 1566 \\
PDT & 181 & 391 & 756 & 1362 \\
IN & 190 & 385 & 760 & 1408 \\
PRP\$ & 189 & 383 & 745 & 1386 \\
DT & 190 & 379 & 734 & 1363 \\
MD & 177 & 378 & 754 & 1392 \\
TO & 182 & 368 & 734 & 1363 \\
\bottomrule
\end{tabular}
\caption{Mean number of steps a token lasts, grouped by part-of-speech tags. \label{tab:full_pos}}
\end{table}